\documentclass{article}

\usepackage[preprint]{corl_2026} 
\usepackage{graphicx}
\usepackage{algorithm}
\usepackage{algpseudocode}
\usepackage{amsmath, amssymb, mathtools}
\usepackage{wrapfig}
\usepackage{cancel}
\usepackage[most]{tcolorbox}
\usepackage[table]{xcolor} 
\usepackage{xcolor}
\usepackage{tikz}
\usetikzlibrary{shapes, arrows.meta, positioning, shadows, calc, fit}
\usepackage{booktabs} 
\usepackage{multirow} 
\usepackage[normalem]{ulem}
\usepackage{flushend}
\usepackage{balance}
\usepackage{makecell}
\usepackage[cal=boondoxo]{mathalfa}
\usepackage{capt-of}  
\usepackage{subcaption}
\usepackage{graphicx}
\usepackage{siunitx}
\usepackage[most]{tcolorbox}
\tcbset{
  promptbox/.style={
    enhanced,
    breakable,
    colback=gray!5,
    colframe=black!15,
    boxrule=0.4pt,
    arc=1.2mm,
    left=1.2mm,right=1.2mm,top=1.0mm,bottom=1.0mm,
  }
}

\usetikzlibrary{positioning,arrows.meta,fit,backgrounds}

\definecolor{objblue}{RGB}{222,235,247}
\definecolor{objborder}{RGB}{88,141,194}
\definecolor{reggreen}{RGB}{226,240,217}
\definecolor{regborder}{RGB}{96,153,102}
\definecolor{terorange}{RGB}{252,228,214}
\definecolor{terborder}{RGB}{210,130,75}
\definecolor{rootgray}{RGB}{245,245,245}

\hypersetup{
  colorlinks=true,
  linkcolor=blue,
  citecolor=blue,
  urlcolor=blue,
  pdftitle={CoFL-S: Spatially Queryable Sector Flow Fields for Local Language-Conditioned Navigation},
  pdfauthor={Haokun~Liu, Zhaoqi~Ma, Yicheng~Chen, Wentao~Zhang, Masaki~Kitagawa, Zicen~Xiong, Jinjie~Li, Moju~Zhao}
  }
\usepackage[figure,figure*,table,table*]{hypcap} 

\definecolor{stageblue}{rgb}{0.05,0.30,0.62}

\algnewcommand{\LineComment}[1]{\Statex \hfill \textcolor{gray!90}{\footnotesize \textit{$\triangleright$ #1}}}

\title{CoFL-S: Spatially Queryable Sector Flow Fields for Local Language-Conditioned Navigation}

\author{
  \textnormal{Haokun~Liu \quad Zhaoqi~Ma \quad Yicheng~Chen} \\
  \textnormal{Wentao~Zhang \quad Masaki~Kitagawa \quad Zicen~Xiong
    \quad Jinjie~Li \quad Moju~Zhao} \\[0.8em]
  \textbf{Dragon Lab, The University of Tokyo}
}

\begin{document}
\maketitle
\vspace{-2em}
\begin{center}
  \includegraphics[width=\linewidth]{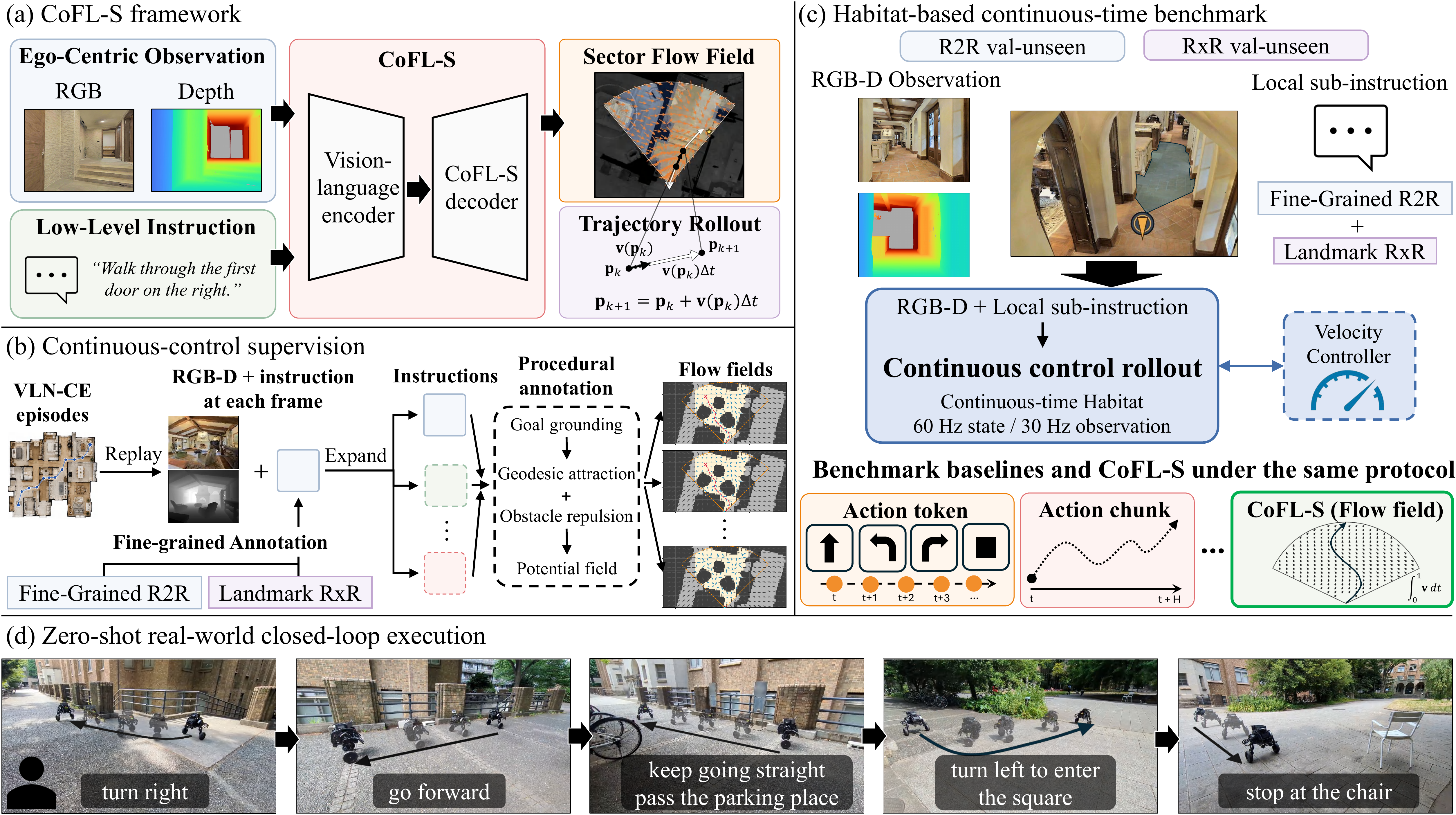}
  \captionof{figure}{{Overview of CoFL-S.} CoFL-S learns an ego-centric sector flow field from RGB-D observations and local language instructions, and extracts continuous control through field rollout. The framework is trained with dense sector flow field supervision and evaluated on a Habitat-based benchmark built from R2R/RxR episodes and zero-shot real-world closed-loop execution.
  }
  \label{fig:overview}
\end{center}
\vspace{-0.3em}

\begin{abstract}
Vision-Language Navigation has increasingly emphasized high-level instruction reasoning, memory, global map construction, and instruction decomposition, while the low-level action representation remains comparatively underexplored. We propose CoFL-S, a low-level vision-language-action framework that predicts a language-conditioned flow field over the robot's local visible sector and generates continuous trajectories by rolling out the predicted field. To train this low-level representation, we convert each VLN-CE episode, originally a whole-episode instruction paired with an action sequence, into frame-level local supervision with aligned sub-instructions and matched action, trajectory, and dense flow-field targets. For evaluation, we introduce a continuous-time Habitat benchmark that isolates low-level action interfaces from instruction decomposition and executes all methods through a shared velocity-command controller, enabling decomposition-independent closed-loop comparison across different planner frequencies rather than fixed discrete forward-and-turn transitions in VLN-CE. Under matched encoders and training settings, CoFL-S consistently outperforms baselines across planner frequencies in the continuous-time Habitat benchmark, and zero-shot real-world closed-loop deployment further shows its advantage over the evaluated baselines beyond simulation. See the project website at: \href{https://github.com/ut-dragon-lab/CoFL}{https://github.com/ut-dragon-lab/CoFL}.
\end{abstract}



\section{Introduction}

Vision-Language Navigation (VLN) is an embodied AI problem that couples high-level semantic understanding, visual perception, and motion grounding: an agent must interpret natural-language instructions, perceive its surroundings, and convert the intended semantic progress into executable motion. Since benchmark settings such as R2R~\cite{anderson2018vln}, RxR~\cite{ku2020rxr}, and VLN-CE~\cite{krantz2020vlnce} established the task, recent progress has substantially strengthened the perception-and-reasoning side of this pipeline through semantic grounding, memory construction, map reconstruction, and long-horizon instruction decomposition~\cite{hong2021vlnbert,chen2021hamt,zhang2025uninavid,cheng2025navila}. However, high-level reasoning only determines what the agent should do next; the execution layer still determines how this intent is translated into robot motion.

This low-level question is difficult to study in standard VLN evaluations, where instruction understanding, progress estimation, and motion grounding are evaluated as one coupled system. As a result, the effect of the action interface itself is often entangled with high-level semantic reasoning. Fine-grained VLN annotations provide a way to expose this question: long instructions can be aligned with local route segments, yielding sub-instructions that describe short-horizon semantic progress~\cite{hong2020sub,he2021landmark}. Given the current observation and such a local sub-instruction, we can ask a focused execution question: what action representation should convert local intent into executable motion?

Existing executable interfaces for VLN and embodied control include discrete actions, waypoints, and action chunks~\cite{krantz2020vlnce,krantz2021waypoint,krantz2022sim2sim,brohan2022rt1,zhao2023act,chi2023diffusion}. These representations are effective and practical, but they typically specify a current-state action, a target, or a finite-horizon sequence of future actions. They therefore do not explicitly describe how the desired motion should vary across nearby workspace locations, obstacle configurations, or replanning frequencies. Classical navigation and collision-avoidance methods suggest a complementary view: local control can be represented as spatially indexed guidance. Potential fields, navigation functions, and velocity-space planners specify how motion changes across workspace locations or velocity states, producing corrective behavior under different offsets and obstacle configurations~\cite{khatib1986real,rimon1992exact,fiorini1998motion,fox2002dynamic}. This motivates a low-level VLN interface that maps local semantic instruction and observation to a spatially queryable control field.

We introduce CoFL-S, a sector flow-field framework for language-conditioned local navigation. Given an ego-centric RGB-D observation and a low-level sub-instruction, CoFL-S predicts a language-conditioned 2D flow field over the robot’s visible local ground sector, queries it in normalized sector coordinates, and integrates it in a robot-centric ground-plane frame to generate a continuous local trajectory.
To train this representation of low-level motion grounding, we use fine-grained VLN annotations~\cite{hong2020sub,he2021landmark,he2026fine} to align replayed R2R-CE and RxR-CE frames~\cite{krantz2020vlnce} with local sub-instructions. For each aligned frame-instruction pair, we retain the original discrete action label and add matched local trajectory and dense flow-field targets. We further add object- and region-grounded instruction slots to increase local language diversity.
For evaluation, we introduce a continuous-time Habitat benchmark with frame-aligned sub-instruction annotations along each episode's reference path. At each closed-loop step, the agent's position is projected onto the reference path to retrieve the corresponding sub-instruction, which is paired with the current observation and fed to each method. Each method then predicts its own low-level interface---an action token, a finite-horizon action chunk, or a sector flow field---which is converted into velocity commands and executed by a shared controller, enabling decomposition-independent comparison across planner frequencies rather than fixed discrete forward-and-turn transitions in VLN-CE~\cite{krantz2020vlnce}.

\noindent\textbf{Contributions}. Our contributions are threefold. (i) We propose CoFL-S, a spatially queryable sector flow-field representation for first-person local language-conditioned navigation (Fig.~\ref{fig:overview}a; Sec.~\ref{sec:cofl_framework}). (ii) We convert VLN-CE~\cite{krantz2020vlnce} episodes from whole-episode instruction--action traces into frame-level local supervision, pairing each RGB-D frame with an aligned sub-instruction and matched action, trajectory, and flow-field targets for low-level interface comparison (Fig.~\ref{fig:overview}b; Sec.~\ref{sec:dataset}).  (iii) We introduce a Habitat-based sub-instruction-aligned continuous-time benchmark for simulation evaluation. 
CoFL-S consistently outperforms representative baselines across planner frequencies in simulation, and further achieves stronger zero-shot closed-loop performance than the evaluated baselines in real-world indoor, outdoor, and hybrid indoor--outdoor deployments (Fig.~\ref{fig:overview}c,d; Sec.~\ref{sec:exp}).

\section{Related Work}

\subsection{Vision-Language Navigation}

VLN studies how an embodied agent follows natural-language instructions under partial observability.
Canonical benchmarks such as R2R~\cite{anderson2018vln}, RxR~\cite{ku2020rxr}, and VLN-CE~\cite{krantz2020vlnce} established instruction following in photo-realistic indoor environments and continuous simulation~\cite{mp3d,savva2019habitat}.
Recent VLN methods have advanced instruction grounding, mapping, subgoal prediction, memory, and instruction decomposition, while also examining visual and physical embodiment gaps~\cite{hong2021vlnbert,chen2021hamt,zhang2025uninavid,cheng2025navila,Han_2025_CVPR,Yao_2025_ICCV,hong2025general,Wang_2025_ICCV}.
This paper takes a complementary perspective: using fine-grained annotations from Fine-Grained R2R~\cite{hong2020sub} and Landmark-RxR~\cite{he2021landmark,he2026fine}, it factors out high-level route decomposition and studies how local language-conditioned intent should be represented for low-level execution.

\subsection{Action Representations and Spatial Guidance}

VLN commonly uses discrete navigation actions such as moving forward, turning, and stopping, or predicts relative waypoint targets that are typically realized by an additional tracking or local-planning module~\cite{krantz2020vlnce,zhang2025uninavid,krantz2021waypoint,krantz2022sim2sim}.
Beyond navigation, robot policies often use action-token policies~\cite{brohan2022rt1,rt2,openvla} or action-chunk decoders~\cite{zhao2023act,chi2023diffusion,pi0,pi0.5}.
This distinction guides our baseline choice: action tokens represent the current-state decision interface, while action chunks provide an end-to-end counterpart to waypoint interfaces by predicting both a local target and the motion rollout.
Together, these baselines cover current-state decisions and finite-horizon rollouts, but neither represents how desired motion should vary across nearby workspace locations.
Classical navigation methods offer a complementary spatial-guidance view, representing control over workspace locations or velocity states rather than only at the current robot state~\cite{khatib1986real,rimon1992exact,fiorini1998motion,fox2002dynamic,dijkstra1959note,hart1968formal}.
A recent learned flow-field policy~\cite{liu2026cofl} connects perception, semantics, and control by representing motion as a flow field over a globally observed bird's-eye view (BEV) workspace, enabling dense workspace-level supervision and stronger navigation feedback than action or trajectory interfaces.
However, its third-person BEV observation assumption bypasses the partial-observability and onboard-sensing constraints faced by robots that must infer control from first-person sensory input.
CoFL-S brings this principle to a more restrictive and practical setting, predicting a language-conditioned visible-sector field from ego-centric RGB-D observations for onboard closed-loop execution.

\section{CoFL-S Framework}
\label{sec:cofl_framework}

\subsection{Problem Formulation and Sector Flow Field Representation}

CoFL-S is designed as a low-level planner for the visible local workspace within the robot camera's field of view. 
Given an ego-centric RGB observation $I$, depth $D$, and a low-level language instruction $\ell$, the desired output is a local continuous trajectory $\tau(t)\in\mathbb R^2$ parameterized by normalized time $t\in[0,1]$.
The trajectory is represented in the robot-centric ground-plane Cartesian frame as
\begin{equation}
  \tau(t)=(x_{\mathrm{fwd}}(t),y_{\mathrm{lft}}(t)),\quad\tau(0)=(0,0).
\end{equation}

\begin{figure}[t]
  \includegraphics[width=\linewidth]{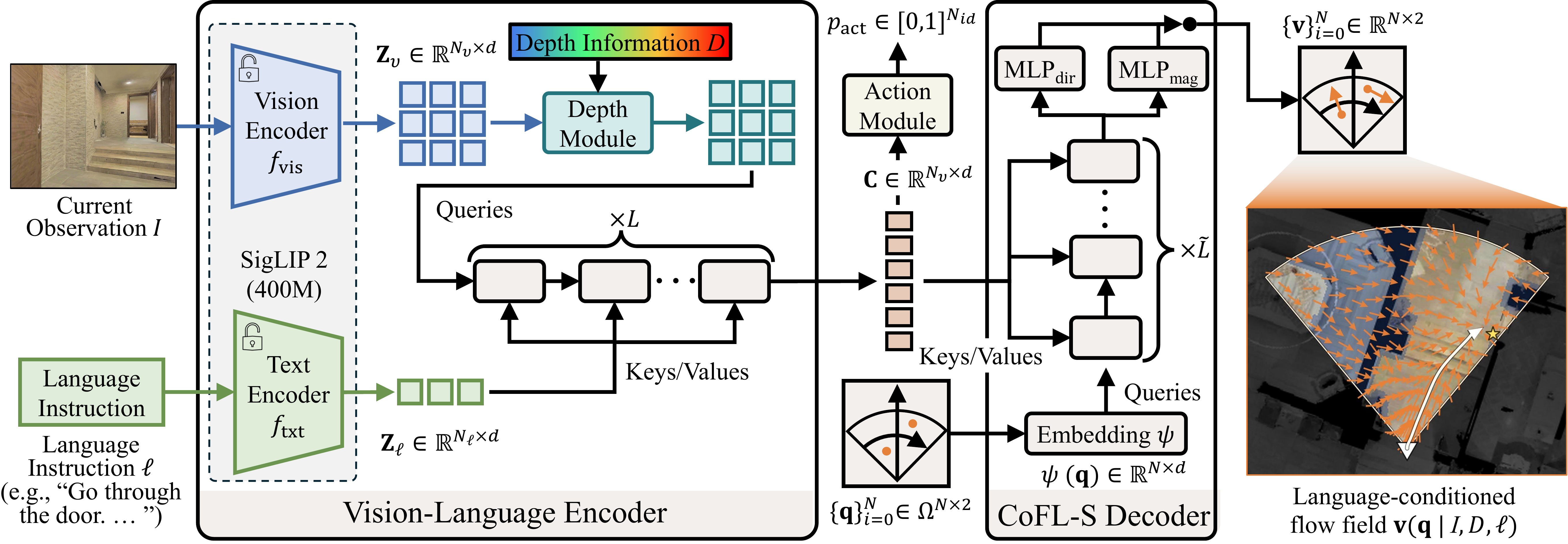}
  \vspace{-1.5em}
  \caption{Overview of CoFL-S architecture. A vision-language encoder and query-based sector decoder map RGB-D observations and local instructions to a spatially queryable sector flow field.}
  \label{fig:cofl_arch}
\end{figure}

Instead of regressing the whole trajectory, CoFL-S generates it through a first-order motion model:
\begin{equation}
  \dot{\tau}(t)
  =
  \mathbf v_{\phi}
  \left(
  \Pi_{\mathrm{polar}}(\tau(t))
  \mid I,D,\ell
  \right).
  \label{eq:cofls_traj_ode}
\end{equation}

Here $\mathbf v_{\phi}$ is a learnable flow field conditioned on the current observation and instruction, and $\Pi_{\mathrm{polar}}$ maps a Cartesian point $\tau(t)$ on the local ground plane to its normalized sector coordinate $\mathbf q$.
We define the normalized polar sector as $\Omega=[-1,1]\times[0,1]$ and $\mathbf q=(\tilde{\theta},\tilde r)\in\Omega$. This polar sector is naturally aligned with the robot camera's field of view.

For a queried sector coordinate, the policy predicts a Cartesian velocity:
\begin{equation}
  \mathbf v(\mathbf q\mid I,D,\ell)
  =
  (v_{\mathrm{fwd}},v_{\mathrm{lft}})
  \in\mathbb R^2.
\end{equation}

Thus, the field is queried in normalized polar coordinates, while both the predicted velocity and the generated trajectory are expressed in the ego-centric ground-plane Cartesian frame.

\subsection{CoFL-S Architecture}
\label{subsec:cofl_archi}

\begin{wrapfigure}[21]{r}{0.41\linewidth}
  \centering
  \vspace{-1.8em}

  \begin{subfigure}{\linewidth}
    \centering
    \includegraphics[width=\linewidth]{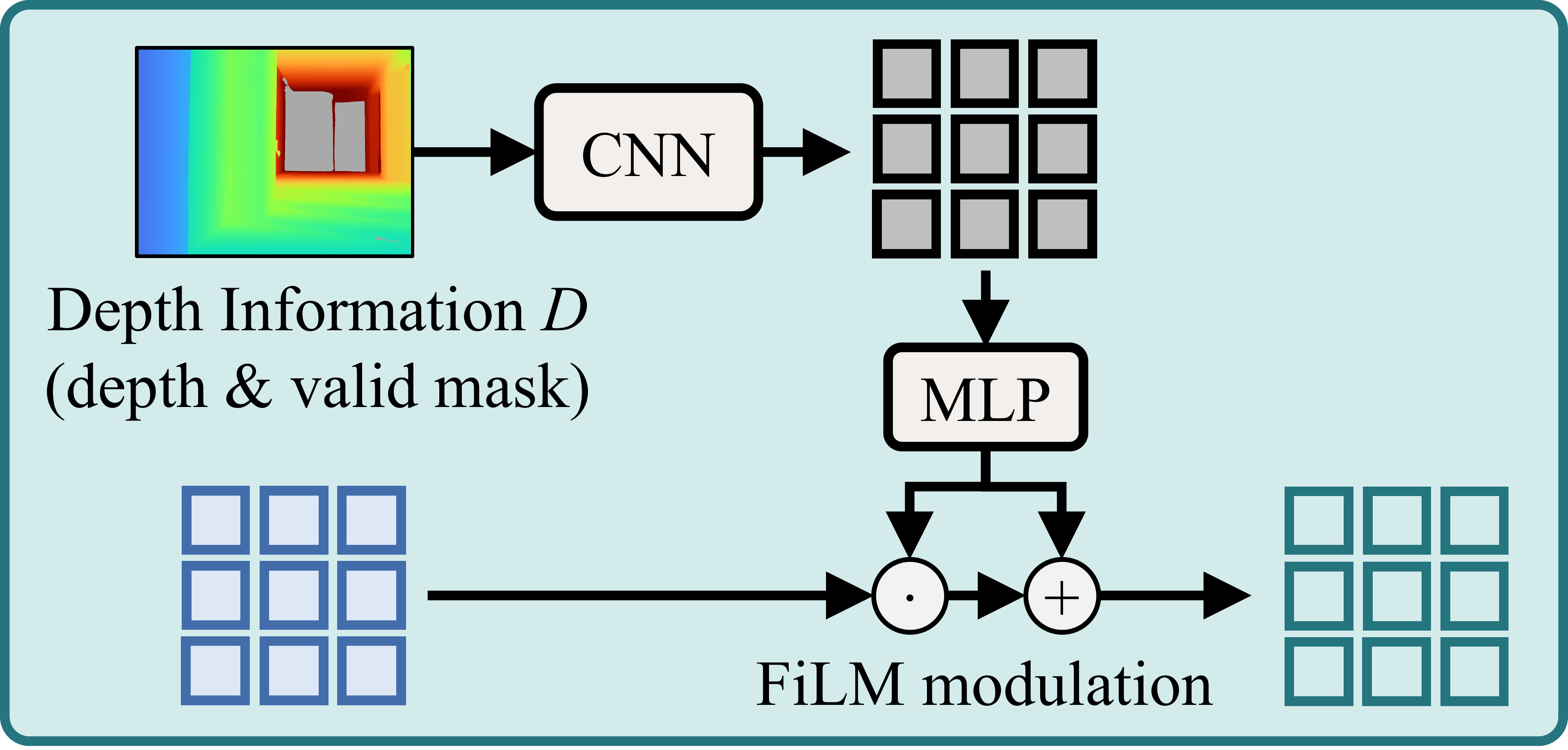}
    \vspace{-1.5em}
    \caption{Depth module}
    \label{fig:cofl_depth}
  \end{subfigure}
  \par
  \begin{subfigure}{\linewidth}
    \centering
    \includegraphics[width=\linewidth]{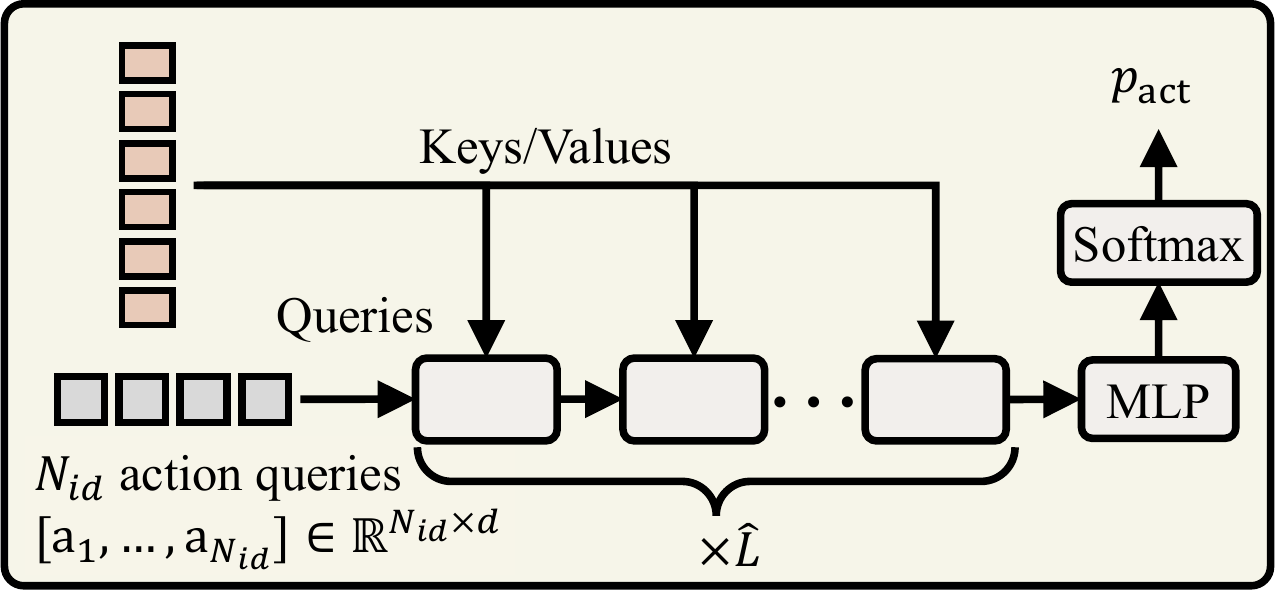}
    \vspace{-1.4em}
    \caption{Action module}
    \label{fig:cofl_action}
  \end{subfigure}

  \vspace{-0.5em}
  \caption{
  Auxiliary modules in CoFL-S.
  The depth module injects geometric information into visual tokens, while the action module uses action queries to predict a discrete termination probability.
  }
  \label{fig:cofl_aux_modules}
  \vspace{-1.0em}
\end{wrapfigure}

Fig.~\ref{fig:cofl_arch} summarizes the architecture. 
An unfrozen SigLIP 2 encoder \cite{tschannen2025siglip2} $(f_{\mathrm{vis}},f_{\mathrm{txt}})$ first maps the RGB observation $I$ and low-level instruction $\ell$ into vision tokens $\mathbf Z_v$ and text tokens $\mathbf Z_\ell$ in a shared embedding space:

\begin{equation}
  \mathbf Z_v =f_{\mathrm{vis}}(I),\quad \mathbf Z_\ell=f_{\mathrm{txt}}(\ell).
\end{equation}

In the RGB-D setting, a convolution $f_{\mathrm{dep}}$ first patchifies the depth into features aligned with the vision tokens by matching the kernel size and stride. Then, a FiLM pathway $f_{\mathrm{FiLM}}$~\cite{perez2018film} injects these depth features into $\mathbf Z_v$ through bounded residual modulation (illustrated in Fig.~\ref{fig:cofl_depth}).
The modulated visual tokens and text tokens are then fused by Transformer decoder-style blocks $f_{\mathrm{fuse}}$ \cite{vaswani2017attention}, where visual tokens serve as queries and text tokens serve as keys and values:

\begin{equation}
  \mathbf C
  =
  f_{\mathrm{fuse}}(f_{\mathrm{FiLM}}(\mathbf Z_v, f_{\mathrm{dep}}(D)),\mathbf Z_\ell).
\end{equation}

For field prediction, each sector query $\mathbf q\in \Omega$ is embedded as a coordinate token with Gaussian Fourier feature embedding $\psi$ followed by a linear projection \cite{tancik2020fourier}. 
A query-based Transformer decoder refines these coordinate tokens over the fused context $\mathbf C$:
\begin{equation}
  \mathbf v(\mathbf q\mid I,D,\ell)
  =
  f_{\mathrm{dec}}( \psi(\mathbf q),\mathbf C).
\end{equation}

In addition to the sector-field decoder, CoFL-S includes a lightweight action module for termination prediction, as illustrated in Fig.~\ref{fig:cofl_action}. 
This module shares the fused context $\mathbf C$ with the decoder and uses the same query-based Transformer-decoder design~\cite{vaswani2017attention}, but replaces coordinate queries with $N_{id}$ learnable action queries $[\mathrm a_1,\ldots,\mathrm a_{N_{id}}]$, corresponding to the discrete VLN-CE action IDs. 

The decoder output is passed to an MLP that produces one logit for each action query:
\begin{equation}
  \mathbf s
  =
  f_{\mathrm{act}}\big([\mathrm a_1,\ldots,\mathrm a_{N_{id}}],\mathbf C\big),
  \qquad
  p_{\mathrm{act}}=\mathrm{Softmax}(\mathbf s).
\end{equation}

\subsection{Learning Objective and Inference}
During training, we sample sector queries $\mathbf q_i \in \Omega$ from an area-uniform distribution over the physical local sector, map them to Cartesian positions $\mathbf p_i$, and bilinearly sample the target Cartesian velocity $\mathbf v_i^*$ from the annotated dense flow field. 
The training objective is
\begin{equation}
  \mathcal L = \mathcal L_{\mathrm{dir}}(\mathbf v(\mathbf q), \mathbf v^*(\mathbf q)) + \lambda_\mathrm{mag}\mathcal L_{\mathrm{mag}}(\mathbf v(\mathbf q), \mathbf v^*(\mathbf q)) +
  \lambda_\mathrm{act}\mathcal L_{\mathrm{act}}(p_\mathrm{act}, p_\mathrm{act}^*).
\end{equation}
Here $\mathcal L_{\mathrm{dir}}$ supervises the local motion direction at each queried position, indicating where the robot should move under the current instruction. 
$\mathcal L_{\mathrm{mag}}$ supervises the distance-to-go scale, which reflects the remaining obstacle-aware geodesic distance to the instruction-specific goal. $\mathcal L_{\mathrm{act}}$ supervises the auxiliary action module used for termination. $p_\mathrm{act}^*$ is a one-hot distribution derived from the ground truth action id $\mathrm a^{*}$.

At each inference cycle, if the STOP action has the highest probability in $p_{\mathrm{act}}$, execution terminates; otherwise, motion is generated by integrating the predicted sector field.
For efficiency, we query the decoder once on a regular polar grid and cache the resulting field; rollout then samples velocities from this grid by bilinear interpolation instead of repeatedly invoking the decoder.
Rollout is performed in Cartesian space with $T$ inference steps. 
Starting from $\mathbf p_0=\tau(0)$, we repeatedly map the current Cartesian state to a sector query and integrate the predicted Cartesian velocity:
\begin{equation}
  \mathbf p_{j+1}
    =
    \mathbf p_j
    +
    \Delta t\,\tilde{\mathbf v}_j,
    \qquad
    \Delta t=\frac{1}{T},
\end{equation}
where $\tilde{\mathbf v}_j=s(j,T)\,{\mathbf v}(\mathbf q_j),\quad \mathbf q_j=\Pi_{\mathrm{polar}}(\mathbf p_j)$.
Here ${\mathbf v}(\mathbf q_j)$ denotes the bilinearly interpolated velocity from the cached sector grid.
$s(j,T)$ is a bounded inverse-time rescaling that converts distance-to-go-scaled vectors into fixed-horizon rollout steps.
The final output is the Cartesian trajectory $\tau=\{\mathbf p_0,\mathbf p_1,\ldots,\mathbf p_T\}$.
Appendix~\ref{app:cofls_framework} gives thorough implementation details.

\begin{figure}[b]
  \includegraphics[width=\linewidth]{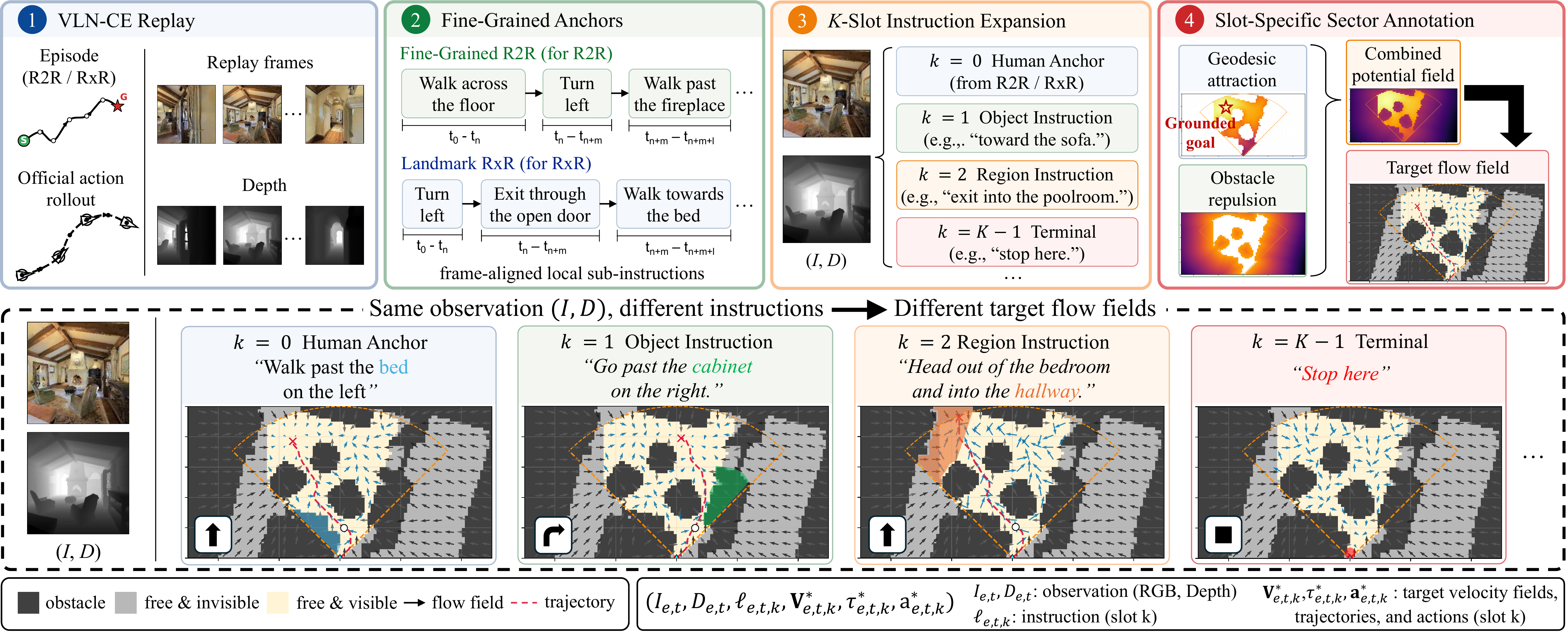}
  \caption{
  Overview of the continuous-control supervision augmentation. 
  Each replayed VLN-CE frame is aligned with a fine-grained human anchor and expanded into $K$ local instruction slots. The slots share the same RGB-D observation but induce different local goals and target trajectories and flow fields, generated by combining geodesic attraction with obstacle/invisible-region repulsion.
  }
  \label{fig:data_aug}
  \vspace{-0.4em}
\end{figure}

\section{Continuous-Control Supervision from VLN-CE Replay}
\label{sec:dataset}
VLN-CE episodes~\cite{krantz2020vlnce} built from R2R~\cite{anderson2018vln} and RxR~\cite{ku2020rxr} provide whole-episode supervision mainly in the form of discrete navigation actions.
This is suitable for action-token policies, but does not directly provide the trajectory-level or spatially indexed targets needed by continuous action interfaces.
We therefore convert replayed VLN-CE episodes into frame-level local supervision, where each ego-centric observation is paired with a local sub-instruction and three matched targets: a discrete action label, a short local trajectory, and a dense sector flow field.
Existing fine-grained annotations (Fine-Grained R2R~\cite{hong2020sub} and Landmark-RxR~\cite{he2021landmark}) provide the local sub-instruction anchors, while additional semantically grounded alternatives from visible objects and local regions are used to increase instruction diversity.
During Habitat~\cite{savva2019habitat} replay, we also record agent poses, the navigation mesh, and simulator semantic observations for offline annotation: the navigation mesh defines reachable free space, while semantic observations identify visible objects and regions used for alternative instruction grounding.

Fig.~\ref{fig:data_aug} summarizes the pipeline.
For each replayed episode $e$ and frame $t$, we construct $K$ instruction-conditioned slots:
\begin{equation}
  \mathbb D_{e,t} =
  \{(I_{e,t},D_{e,t},\ell_{e,t,k},\mathbf V_{e,t,k}^{*},\tau^{*}_{e,t,k},\mathrm a^{*}_{e,t,k})\}_{k=0}^{K-1},
\end{equation}
where $\ell_{e,t,0}$ is the human anchor from fine-grained annotations and $\ell_{e,t,k>0}$ are procedurally generated alternatives that expand the dataset.
All slots share the same observation $(I_{e,t},D_{e,t})$ but correspond to instruction-specific flow fields, trajectories, and action labels.
For each frame-slot pair, we ground the instruction to a reachable local goal and construct an instruction-specific potential field on an ego-centric ground-plane raster. This potential combines cost-weighted geodesic attraction toward the grounded goal~\cite{dijkstra1959note} with distance-transform-based repulsion~\cite{khatib1986real} from obstacles, invisible regions, and local boundaries. Taking the negative gradient of this composed potential gives a dense Cartesian flow field, which is sampled at sector query locations for CoFL-S supervision.
The local trajectory target is extracted from the same geodesic predecessor map, and the discrete action target is obtained by bucketing the trajectory.
Implementation details are provided in Appendix~\ref{app:data_aug_details}.

\section{Experiments}
\label{sec:exp}
We evaluate CoFL-S in two settings: 
a Habitat-based sub-instruction-aligned continuous-time benchmark built on R2R~\cite{anderson2018vln}/RxR~\cite{ku2020rxr} navigation episodes, and zero-shot closed-loop deployment on a real robot.
This section focuses on two main questions: 
(i) How does CoFL-S compare with RGB-D-to-control interfaces under matched encoders and training settings (Sec.~\ref{subsec:main_benchmark})? 
(ii) Can CoFL-S better zero-shot transfer to real-world navigation and run fully onboard in closed-loop execution (Sec.~\ref{subsec:real_world})? 
We also summarize ablations in Sec.~\ref{subsec:ablation} and integrate CoFL-S into a large pretrained VLN model in Sec.~\ref{subsec:largevln}.

\subsection{Benchmark Setup}

\begin{wrapfigure}[7]{r}{0.38\linewidth}
  \vspace{-5.5em}
  \centering
  \includegraphics[width=\linewidth]{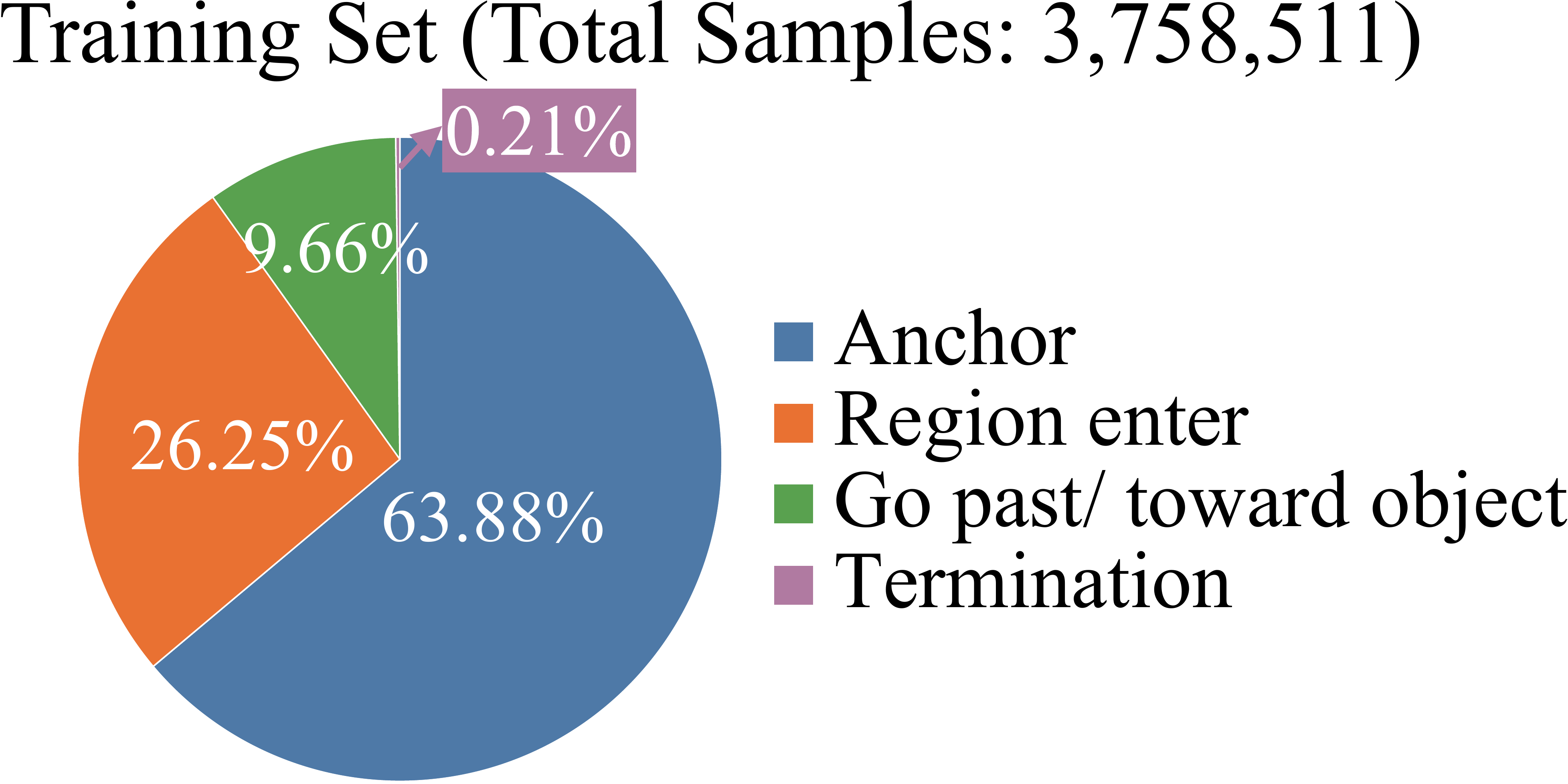}
  \vspace{-1.5em}
  \caption{
    Distribution of instruction types in the training set. 
    Besides the original sub-instructions, the data includes object, region, and minimal terminal variants.
  }
  \label{fig:dataset_distr}
  \vspace{-1.0em}
\end{wrapfigure}

\textbf{Simulation protocol}. Training data are constructed from the official R2R-CE and RxR-CE~\cite{krantz2020vlnce} training episodes by deriving interface-matched discrete, finite-horizon, and dense flow-field targets from the same replayed frames and adding the instruction-diverse slots described in Sec.~\ref{sec:dataset}. Fig.~\ref{fig:dataset_distr} summarizes the composition of the augmented training set. 
We use a subset of the official R2R~\cite{anderson2018vln} and RxR~\cite{ku2020rxr} val-unseen episodes for evaluation, but do not follow the standard discrete VLN-CE action protocol.
Instead, all methods are evaluated in a continuous-time Habitat environment~\cite{savva2019habitat}, where the agent state is updated at \qty{60}{\hertz}, RGB-D observations are received at \qty{30}{\hertz}, and planner outputs are converted into velocity commands executed by a shared low-level controller.
At each control step, we map the current Habitat pose of the agent to the closest progress point on the reference trajectory and select the associated Fine-Grained R2R~\cite{hong2020sub} or Landmark-RxR~\cite{he2021landmark} sub-instruction. 
All methods receive the same ego-centric RGB-D observation stream and aligned local sub-instruction, and are evaluated with the same controller and evaluation episodes.
Inference latency is measured on an NVIDIA GeForce RTX 4090.
More details are provided in Appendix~\ref{app:implementation_details}.

\textbf{Baselines}. All methods use the same vision-language encoder (Sec.~\ref{subsec:cofl_archi}) and matched training.
\textbf{Action Token}, representing discrete action-class prediction, follows the standard VLN-CE discrete action interface, predicting forward, turn-left, turn-right, and stop actions~\cite{krantz2020vlnce,zhang2025uninavid}, but uses continuous velocity control instead of discrete jumps. For architectural matching, it uses the same query-based action module as CoFL-S (Sec.~\ref{subsec:cofl_archi}), but uses its predicted action distribution for all low-level decisions rather than only for termination.
\textbf{Action Chunk}, representing current-state-anchored future-motion prediction, predicts a finite-horizon future action sequence from the current observation and instruction with a Transformer-style diffusion policy head~\cite{chi2023diffusion} (Horizon: $100$, denoise steps: $5$).
\textbf{Waypoint + GT Planner}, representing endpoint prediction followed by privileged geometric planning~\cite{krantz2021waypoint}, uses the same diffusion-head architecture as Action Chunk but denoises only the future trajectory endpoint. Habitat's planner then generates a trajectory to the predicted waypoint using the ground-truth navigation mesh.
\textbf{Velocity Regression}, representing direct continuous velocity prediction~\cite{Blukis-RSS-18}, uses two queries that cross-attend to the encoder context, followed by MLP readouts that predict the forward and lateral velocity components $(v_{\mathrm{fwd}}, v_{\mathrm{lft}})$ in the robot frame.
Action Chunk, Waypoint + GT Planner, and Velocity Regression use the same query-based action module as CoFL-S (Sec.~\ref{subsec:cofl_archi}) for STOP prediction.
\textbf{CoFL-S} generates motion by integrating the predicted ego-centric sector flow field, using a $40 \times 40$ query grid and $100$ rollout steps. More details are provided in Appendix~\ref{app:baseline_details}.

\textbf{Metrics}. We report standard VLN metrics, including \textbf{navigation error} (NE), \textbf{oracle success} (OS), \textbf{success rate} (SR), and \textbf{Success weighted by Path Length} (SPL).
We further report two metrics beyond standard VLN metrics:
\textbf{Blocked-Step Rate} (BSR), computed as the fraction of control steps in which the agent is blocked by the navigation mesh,
and \textbf{Heading Smoothness} (HS), computed from heading changes along the executed trajectory. More details are provided in Appendix~\ref{app:metric_definitions}.

\begin{figure}[t]
  \includegraphics[width=\linewidth]{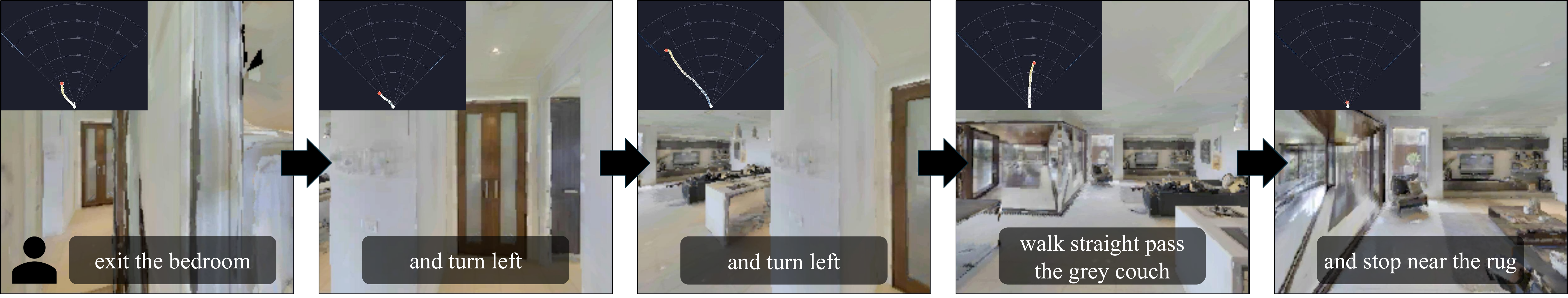}
  \vspace{-1.5em}
  \caption{
    Qualitative snapshots of CoFL-S executing a task sequence in R2R-CE, composed of fine-grained sub-instructions in a Habitat environment~\cite{savva2019habitat}.
    For each step, we visualize the RGB observation, the aligned local instruction (illustration of alignment is shown in Fig.~\ref{fig:topdown_aligned}), and the predicted trajectory integrated from the sector flow field. More examples are provided in Appendix~\ref{app:more_benchmark_example}.
  }
  \label{fig:benchmark}
  \vspace{-0.3em}
\end{figure}

\subsection{Main Results}
\label{subsec:main_benchmark} 
Table~\ref{tab:main_benchmark} compares the action-token, action-chunk, waypoint, velocity regression, and the CoFL-S interface under the same simulation protocol. A qualitative example is provided in Fig.~\ref{fig:benchmark}. Each method is tested at low, moderate, and high frequencies (\qty{2}{\hertz} / \qty{5}{\hertz} / \qty{10}{\hertz}).

\begin{table*}[h]
\centering
\small
\vspace{-0.4em}
\setlength{\tabcolsep}{3.2pt}
\renewcommand{\arraystretch}{1.05}
\caption{Main results on R2R-CE and RxR-CE val-unseen under our continuous-time benchmark.}
\vspace{-0.8em}
\label{tab:main_benchmark}
\resizebox{\linewidth}{!}{
\begin{tabular}{lccccccc|ccccccc}
\toprule
\multirow{2}{*}{Method} & \multirow{2}{*}{Head param (M)} & \multicolumn{6}{c|}{R2R-CE (1,200 episodes)} & \multicolumn{6}{c}{RxR-CE (1,600 episodes)} & \multirow{2}{*}{Lat. (ms)}\\
\cmidrule(lr){3-8}\cmidrule(lr){9-14}
& & NE$\downarrow$ & OS$\uparrow$ & SR$\uparrow$ & SPL$\uparrow$ & BSR$\downarrow$ & HS$\uparrow$
& NE$\downarrow$ & OS$\uparrow$ & SR$\uparrow$ & SPL$\uparrow$ & BSR$\downarrow$ & HS$\uparrow$ \\
\midrule
Action Token (\qty{2}{\hertz})
& 14.92 & \underline{7.02} & 0.30 & 0.23 & 0.18 & {0.22} & 0.87
& \underline{7.76} & 0.25 & 0.17 & 0.15 & {0.24} & 0.88 & 10.71 \\
Action Chunk (\qty{2}{\hertz})
& 81.21 & 7.29 & \underline{0.43} & \underline{0.28} & \underline{0.21} & 0.36 & {0.90}
& 8.09 & \underline{0.41} & \underline{0.23} & {0.19} & 0.37 & {0.89} & 25.71 \\
Waypoint + GT Planner (\qty{2}{\hertz})
& 81.21 & 7.33 & {0.42} & {0.26} & {0.20} & \textbf{0.11} & \underline{0.91}
& 8.16 & {0.40} & {0.23} & \underline{0.21} & \textbf{0.14} & \underline{0.90} & 18.15 \\
Velocity Regression (\qty{2}{\hertz})
& 29.14 & 7.08 & {0.29} & {0.22} & {0.18} & 0.25 & {0.88}
& 7.86 & {0.28} & {0.19} & {0.16} & 0.26 & 0.89 & 20.14 \\
\rowcolor{gray!15}
CoFL-S (Ours, \qty{2}{\hertz})
& 30.28 & \textbf{6.64} & \textbf{0.52} & \textbf{0.37} & \textbf{0.29} & \underline{0.14} & \textbf{0.92}
& \textbf{7.66} & \textbf{0.51} & \textbf{0.33} & \textbf{0.25} & \underline{0.15} & \textbf{0.92} & 26.07 \\

\midrule
Action Token (\qty{5}{\hertz})
& 14.92 & \underline{6.79} & \underline{0.43} & \underline{0.30} & \underline{0.25} & {0.15} & {0.92}
& {7.63} & \underline{0.38} & \underline{0.23} & \underline{0.20} & {0.17} & {0.92} & 10.71 \\

Action Chunk (\qty{5}{\hertz})
& 81.21 & 7.18 & 0.43 & 0.29 & 0.23 & 0.27 & \underline{0.94}
& 8.34 & 0.36 & 0.22 & 0.19 & 0.40 & \underline{0.93} & 25.71  \\

Waypoint + GT Planner (\qty{5}{\hertz})
& 81.21 & 7.25 & {0.41} & {0.27} & {0.23} & \textbf{0.12} & {0.90}
& 8.46 & {0.36} & {0.21} & {0.19} & \textbf{0.12} & 0.92 & 18.15 \\

Velocity Regression (\qty{5}{\hertz})
& 29.14 & 6.96 & {0.42} & {0.30} & {0.24} & 0.17 & {0.91}
& \underline{7.61} & 0.37 & 0.21 & 0.20 & 0.37 & 0.92 & 20.14 \\

\rowcolor{gray!15}
CoFL-S (Ours, \qty{5}{\hertz})
& 30.28 & \textbf{6.31} & \textbf{0.50} & \textbf{0.37} & \textbf{0.30} & \underline{0.14} & \textbf{0.95}
& \textbf{7.60} & \textbf{0.49} & \textbf{0.33} & \textbf{0.26} & \underline{0.16} & \textbf{0.96} & 26.07\\
\midrule
Action Token (\qty{10}{\hertz})
& 14.92 & \underline{6.67} & \underline{0.42} & \underline{0.31} & \underline{0.25} & \textbf{0.10} & \textbf{0.98}
& {7.65} & {0.39} & {0.27} & {0.19} & \underline{0.11} & \textbf{0.98} & 10.71 \\

Action Chunk (\qty{10}{\hertz})
& 81.21 & 7.33 & 0.39 & 0.27 & 0.21 & 0.22 & \underline{0.97}
& 8.55 & {0.33} & 0.19 & 0.17 & 0.22 & 0.94 & 25.71 \\

Waypoint + GT Planner (\qty{10}{\hertz})
& 81.21 & 7.30 & {0.38} & {0.27} & {0.22} & \underline{0.11} & {0.90}
& 8.46 & {0.33} & {0.21} & {0.19} & \textbf{0.10} & {0.95} & 18.15 \\

Velocity Regression (\qty{10}{\hertz})
& 29.14 & 6.81 & {0.41} & {0.30} & {0.23} & 0.15 & {0.90}
& \underline{7.63} & \underline{0.40} & \underline{0.27} & \underline{0.20} & 0.17 & {0.96} & 20.14 \\

\rowcolor{gray!15}
CoFL-S (Ours, \qty{10}{\hertz})
& 30.28 & \textbf{6.34} & \textbf{0.48} & \textbf{0.36} & \textbf{0.30} & {0.15} & {0.96}
& \textbf{7.28} & \textbf{0.49} & \textbf{0.34} & \textbf{0.26} & {0.16} & \underline{0.97} & 26.07 \\
\bottomrule
\end{tabular}}
\vspace{-0.7em}
\end{table*}

The baselines reveal distinct trade-offs in what the policy learns. Action Token and Velocity Regression both predict an immediate action at the current state and exhibit similar performance, particularly struggling at low replanning frequencies. Action Chunk and Waypoint instead extend guidance beyond the current decision: the former learns a motion sequence, whereas the latter learns where to go and delegates execution to a ground-truth planner. However, predicted sequences may lack geometric consistency, while navmesh shortest paths can hug walls, limiting landmark visibility and hindering subsequent instruction grounding.

CoFL-S learns how to move across queried local states through dense workspace-level supervision, providing spatially distributed corrective guidance rather than a single action, sequence, or destination. Its consistently stronger navigation performance, especially compared with immediate-action policies at low frequencies, supports this representation--supervision perspective~\cite{liu2026cofl}. Although specialized baselines can achieve better local safety metrics, CoFL-S combines competitive local control with stronger task completion across frequencies and datasets.

\subsection{Ablation Studies}
\label{subsec:ablation}
Detailed ablation studies are provided in Appendix~\ref{app:abla}.
(i) The supervision-coverage ablation corroborates the benefit of workspace-level field supervision previously demonstrated in CoFL~\cite{liu2026cofl}, extending this finding to ego-centric sector fields (Appendix~\ref{app:coverage_abla}).
(ii) The instruction-diverse data ablation suggests that instruction-diverse supervision provides moderate gains (Appendix~\ref{app:data_abla}). 
(iii) The modulation ablation shows that the action module and depth module improve termination prediction and geometry-aware execution, respectively (Appendix~\ref{app:module_abla}).

\subsection{Real-World Zero-Shot Validation}
\label{subsec:real_world}

\begin{figure}[t]
  \includegraphics[width=\linewidth]{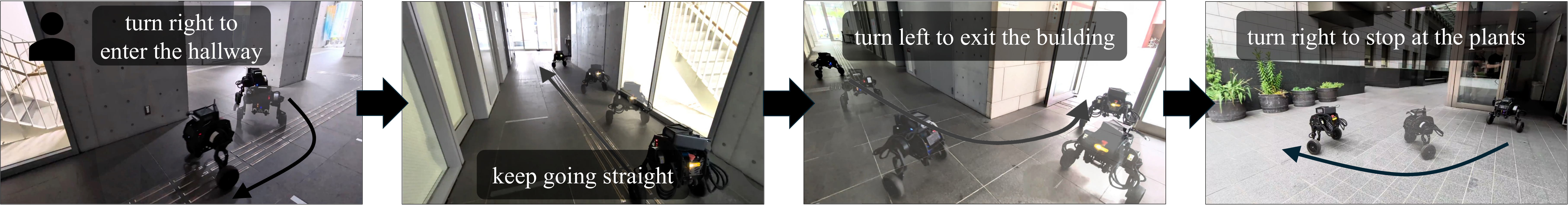}
  \vspace{-1.5em}
  \caption{
    Qualitative snapshots of CoFL-S executing a human-assigned task sequence in a real-world teaching building (hybrid setting). 
    For each step, we visualize the assigned sub-instruction. More examples are provided in Appendix~\ref{app:more_rw_example} (Fig.~\ref{fig:realworld_base} and Fig.~\ref{fig:realworld_tb_1}).
  }
  \label{fig:realworld}
  \vspace{-0.5em}
\end{figure}

We further deploy CoFL-S on a physical mobile robot equipped with an ego-centric Intel RealSense D435i camera (Frequency: \qty{30}{\hertz}; FoV: $70^{\circ}$) and an NVIDIA Jetson AGX Orin \qty{64}{\giga\byte} (details are provided in Appendix~\ref{app:rw_hardware}).
Notably, we directly use the simulation-trained model without any real-world fine-tuning.
The real-world evaluation follows the same local-instruction setting: local subtasks are manually assigned and updated by an operator according to the robot's progress. We evaluate zero-shot deployment across eight task sequences spanning three real-world areas, including indoor, outdoor, and hybrid indoor--outdoor settings. Each sequence is repeated three times, and the inference frequency is fixed at \qty{5}{\hertz}. Target objects are chosen from the Matterport3D label list~\cite{mp3d}. One example is shown in Fig.~\ref{fig:realworld}. The maximum linear velocity is set to \qty{0.5}{\metre\per\second} and the maximum angular velocity is set to \qty{1.0}{\radian\per\second}.

\begin{wraptable}[5]{r}{0.51\linewidth}
\centering
\vspace{-2.3em}
\setlength{\tabcolsep}{4pt}
\caption{
Real-world zero-shot evaluation under human-assigned local instructions.
}
\vspace{-0.7em}
\label{tab:real_world}
\resizebox{\linewidth}{!}{
\begin{tabular}{lccccc}
\toprule
Method & SR$\uparrow$ & CT$\downarrow$ & T-S (s) $\downarrow$ & PL-S (m) $\downarrow$ & Lat. (ms)\\
\midrule
Action Token & 0.46 & 0.63 & 152.17 & 46.54 & 93.72 \\
Action Chunk & 0.33 & 1.42 & 166.91 & 55.42 & 100.58 \\
\rowcolor{gray!15}
CoFL-S       & \textbf{0.75} & \textbf{0.42} & \textbf{111.50} & \textbf{42.43} & {143.29} \\
\bottomrule
\end{tabular}}
\vspace{-0.7em}
\end{wraptable}

Table~\ref{tab:real_world} reports \textbf{success rate} (SR), \textbf{manually counted collision counts averaged over each trial} (CT), \textbf{success-conditioned completion time} (T-S), \textbf{success-conditioned path length} (PL-S) and \textbf{inference latency} (Lat.). 
Under this zero-shot cross-domain deployment setting, CoFL-S achieves the highest SR and the lowest CT, T-S, and PL-S, indicating stable transfer beyond the indoor simulation distribution. 
Compared with Action Token, which remains usable in open areas but suffers from oscillations near obstacles, likely because its current-state decisions are more sensitive to the onboard latency gap (\qty{\sim100}{\milli\second} vs. \qty{\sim10}{\milli\second} in simulation), and Action Chunk, which more often collides or gets stuck, CoFL-S better couples local instruction semantics with surrounding geometric constraints. 
Despite its higher latency, CoFL-S remains compatible with the \qty{5}{\hertz} onboard closed-loop setting, as its rollout provides a short-horizon geometric command that is less sensitive to moderate observation--control latency than single-step action prediction.
Failure analyses are provided in Appendix~\ref{app:failure_analysis}.

\subsection{Integrating CoFL-S with StreamVLN}
\label{subsec:largevln}
\paragraph{Architecture.}
We adapt StreamVLN~\cite{wei2025streamvln} to predict continuous flow fields. We retain its pretrained vision-language backbone and causal input organization, comprising instructions, historical RGB observations, and executed-action tokens, with 32-action windows and eight long-term memory frames. From the multimodal backbone, we extract 196 hidden states corresponding to the current image and the final assistant-prefill state. These form a $197 \times 3584$ context, directly supplied to a two-layer CoFL-S decoder without channel compression. The original vocabulary head is retained on the final prefill state for auxiliary action supervision and online STOP prediction.

\paragraph{Fine-tuning.}
We construct 571,546 training decisions from our R2R/RxR training assets, reserving separate scenes for internal validation. Training uses full instructions. We jointly optimize the visual encoder, multimodal projector, language backbone, field decoder, and vocabulary head. Each decision supervises 1,000 spatial queries using direction and magnitude loss, together with $0.1$-weighted full-vocabulary cross-entropy for the current action. We use BF16, AdamW, a global batch size of 60, and learning rates of $5{\times}10^{-7}$ for vision, $2{\times}10^{-6}$ for the language backbone and native action head, and $5{\times}10^{-5}$ for the field decoder.

\paragraph{Execution and evaluation.}
To accommodate StreamVLN's action-indexed visual history, we use event-triggered replanning rather than fixed-frequency model calls in Sec.~\ref{subsec:main_benchmark}. Each predicted field is integrated into a reference trajectory tracked by the continuous controller. Actual translation and rotation are quantized into forward/left/right history tokens. Replanning occurs after at most four committed tokens or upon trajectory exhaustion. This follows StreamVLN's action-segment organization while accommodating continuous execution. STOP is selected by the native head.
We evaluate both full-instruction and oracle sub-instruction inputs on 1,200 R2R-CE val-unseen episodes (Table~\ref{tab:streamvln_cofl}), using the same R2R-CE evaluation cohort as in Table~\ref{tab:main_benchmark}.

\begin{wrapfigure}{r}{0.5\linewidth}
    \centering
    \vspace{-2.0em}
    \setlength{\tabcolsep}{4pt}
    \captionof{table}{
        StreamVLN and StreamVLN with CoFL-S under full-instruction
        and oracle sub-instruction inputs on our unified benchmark.
    }
    \label{tab:streamvln_cofl}
    \vspace{-0.5em}
    \resizebox{\linewidth}{!}{
        \begin{tabular}{lccccc}
            \toprule
            Method & NE$\downarrow$ & OS$\uparrow$ & SR$\uparrow$
                   & SPL$\uparrow$ & BSR$\downarrow$ \\
            \midrule
            \multicolumn{6}{c}{Full instruction} \\
            \midrule
            StreamVLN          & \textbf{4.71} & 0.65 & 0.57 & 0.52 & 0.07 \\
            \rowcolor{gray!15}
            StreamVLN + CoFL-S & 4.75 & \textbf{0.67} & \textbf{0.58} & \textbf{0.54} & \textbf{0.02} \\
            \midrule
            \multicolumn{6}{c}{Oracle sub-instruction} \\
            \midrule
            StreamVLN          & 4.44 & 0.67 & 0.59 & 0.54 & 0.06 \\
            \rowcolor{gray!15}
            StreamVLN + CoFL-S & \textbf{4.35} & \textbf{0.68} & 0.59 & \textbf{0.56} & \textbf{0.02} \\
            \bottomrule
        \end{tabular}
    }
    \vspace{-1em}
\end{wrapfigure}

\paragraph{Discussion.}
With fine-tuning on a limited set of scenes, CoFL-S integrates into an established VLN framework, extending discrete action-token outputs to continuous trajectory generation through spatially queryable fields. Despite imperfect adaptation, it reduces collision rates measured by BSR while maintaining navigation performance. Remaining mismatches in quantized action history and observation timing motivate memory inputs tailored to continuous navigation.


\section{Conclusion}
\label{sec:conclusion}

We present CoFL-S, a spatially queryable sector flow-field framework that converts local instructions and ego-centric RGB-D observations into continuous trajectories through visible-sector field rollout.
To train and evaluate this representation, we constructed sub-instruction-aligned continuous-control supervision and a Habitat-based sub-instruction-aligned continuous-time benchmark that compares action-token, action-chunk, waypoint, velocity regression, and flow-field interfaces under matched observations, instructions, encoders, and controllers.
Across planner frequencies, CoFL-S achieves stronger task-completion performance than the evaluated baselines in simulation, and zero-shot real-world deployment further shows improved closed-loop navigation over two selected representative baselines.
These results suggest that flow-field policies with dense workspace-level supervision effectively provide numerous corrective motion targets rather than a current-state action or one rollout per training instance.
More fundamentally, when the high-level sub-instruction is fixed, different interfaces yield different results, showing that the low-level action representation is a problem worth studying in VLN.

\section{Limitations and Future Work}

(i) CoFL-S focuses on low-level instruction-to-action control and assumes that a local sub-instruction is available. It does not solve long-horizon route decomposition, exploration, or persistent memory by itself. 
A natural direction is to pair CoFL-S with a high-level task decomposer while using the sector field as the execution layer. 
(ii) The current field is defined on a two-dimensional camera-centered ground-plane sector. This is effective for ground steering and obstacle-aware motion but does not explicitly model full 3D collision avoidance or vertical clearance.
Extending the representation toward 3D spatial fields is an important future direction.

\acknowledgments{This work was supported by JST ASPIRE, Grant Number JPMJAP2521. We thank the anonymous reviewers and the area chair for their constructive feedback.}

\bibliography{reference}

\clearpage
\appendix

\section{Notation and Tensor Shapes}

Table~\ref{tab:cofls_notation} summarizes the symbols used in this paper. 

\begin{table}[h]
\centering
\small
\setlength{\tabcolsep}{4pt}
\renewcommand{\arraystretch}{1.08}
\caption{Notation used in this paper.}
\label{tab:cofls_notation}
\resizebox{\linewidth}{!}{
\begin{tabular}{lll}
\toprule
\textbf{Symbol} & \textbf{Domain} & \textbf{Description} \\
\midrule
\multicolumn{3}{c}{\textit{Problem Setup}} \\
\midrule
$\Omega$ & $[-1,1]\times[0,1]$ & Normalized camera-centered polar sector. \\
$I,D$ & $\mathbb R^{H\times W\times 3},\mathbb R^{H\times W}$ & ego-centric RGB and depth observation. \\
$\ell$ & text & Local language command. \\
$\mathbf q$ & $\Omega$ & Normalized polar query $(\tilde\theta,\tilde r)$. \\
$\mathbf p$ & $\mathbb R^2$ & Cartesian position $(x_{\mathrm{fwd}},y_{\mathrm{lft}})$. \\
$\mathbf v_\phi(\cdot\mid I,D,\ell)$ & $\Omega\rightarrow\mathbb R^2$ & learnable conditioned sector flow policy. \\
$\tau$ & $\mathbb R^{(T+1) \times 2}$ & Local rollout trajectory. \\
\midrule
\multicolumn{3}{c}{\textit{Model and Querying}} \\
\midrule
$\mathbf Z_v, \mathbf Z_\ell$ & $\mathbb R^{N_v\times d_{\mathrm{emb}}},\mathbb R^{N_\ell\times d_{\mathrm{emb}}}$ & Unfrozen visual tokens and text tokens. \\
$\tilde{\mathbf Z}_v,\tilde{\mathbf Z}_\ell$ & $\mathbb R^{N_v\times d},\mathbb R^{N_\ell\times d}$ & Projected vision/text tokens. \\
$\mathbf C$ & $\mathbb R^{N_v\times d}$ & Fused action context tokens. \\
$\mathbf X=\{\mathbf q_i\}_{i=1}^{N}$ & $\Omega^{N\times 2}$ & Decoder query coordinates. \\
$\mathbf Q=\psi(\mathbf X)$ & $\mathbb R^{N\times d}$ & Coordinate query tokens. \\
$\mathbf V(\mathbf X)$ & $\mathbb R^{N\times2}$ & Queried Cartesian velocity vectors. \\
$\mathbf A^{(0)}$ & $\mathbb R^{N_{\mathrm{id}}\times d}$ & Learnable action-ID query tokens. \\
$\mathbf s,p_{\mathrm{act}}$ & $\mathbb R^{N_{\mathrm{id}}}$ & Action logits and softmax action probability. \\
\midrule
\multicolumn{3}{c}{\textit{Training and Inference}} \\
\midrule
$\mathbf V_{e,t,k}^{*}(\cdot)$ & $\mathbb R^2$ & Slot-specific target flow field. \\
$\mathrm a^*_{e,t,k},p^*_{\mathrm{act}}$ & $\{1,\ldots,N_{\mathrm{id}}\},\mathbb R^{N_{\mathrm{id}}}$ & Ground-truth action ID and one-hot action target. \\
$N_s$ & $\mathbb Z^+$ & Number of sampled training queries. \\
$\lambda_\text{mag},\lambda_\text{act},\epsilon$ & $\mathbb R^+$ & Loss weights and numerical stabilizer. \\
$\hat{\mathbf V}$ & $\mathbb R^{\tilde N_\theta\times \tilde N_r\times2}$ & Cached inference velocity grid. \\
$c_{\mathrm{stop}}$ & $\{1,\ldots,N_{\mathrm{id}}\}$ & STOP action ID. \\
$T,\Delta t$ & $\mathbb Z^+,\mathbb R^+$ & Rollout horizon and timestep. \\
$\alpha,\beta$ & $\mathbb R^+$ & Fixed inference rescaling parameters. \\
\midrule
\multicolumn{3}{c}{\textit{Augmentation and Annotation}} \\
\midrule
$e,t,k$ & indices & Episode, replay frame, and command slot. \\
$M_{\mathrm{free}},M_{\mathrm{vis}},M_{\mathrm{eff}}$ & $\{0,1\}^{H_b\times W_b}$ & Walkable, depth-visible, and effective free-space masks. \\
$D_g^w,D_g^{\mathrm{pix}}$ & $\mathbb R^+$ & Cost-weighted and geometric distance-to-go fields. \\
$D_{\mathrm{obs}},\Phi_k$ & $\mathbb R^+$ & Invalid-region distance and composed potential. \\
\bottomrule
\end{tabular}
}
\end{table}

\section{CoFL-S Framework Details}
\label{app:cofls_framework}

This appendix provides the complete implementation details of CoFL-S. 
The model consists of a vision--language encoder pipeline $(f_{\mathrm{vis}},f_{\mathrm{txt}},f_{\mathrm{dep}},f_{\mathrm{FiLM}},f_{\mathrm{fuse}})$ that produces command-conditioned context tokens, a sector decoder $f_{\mathrm{dec}}$ that maps normalized sector queries to ego-BEV Cartesian velocity vectors, and an action module $f_{\mathrm{act}}$ that provides termination signal to the execution.
The predicted field is static and spatially queryable; trajectories are derived after field prediction.
Relative to Sec.~\ref{sec:cofl_framework}, this appendix uses the same symbols but lifts the single-query notation $\mathbf q$ to a batched query set $\mathbf X=\{\mathbf q_i\}_{i=1}^{N}$ for implementation details.

\subsection{Polar--Cartesian Coordinate Conversion}
\label{app:polar_cartesian}

CoFL-S uses normalized polar coordinates to query the visible sector, while trajectories and velocities are represented in the ego-centric ground-plane Cartesian frame. 
A normalized sector query is denoted as
\begin{equation}
  \mathbf q=(\tilde{\theta},\tilde r)\in\Omega,
  \qquad
  \Omega=[-1,1]\times[0,1].
\end{equation}
It is first converted to physical polar coordinates by
\begin{equation}
  \theta
  =
  \tilde{\theta}\frac{\mathrm{HFoV}}{2},
  \qquad
  r
  =
  \tilde rz_{\max},
\end{equation}
where $\mathrm{HFoV}$ is the horizontal field of view and $[0,z_{\max}]$ defines the active physical radial range. 
The Cartesian position in the ego-BEV frame is then
\begin{equation}
  \Pi_{\mathrm{cart}}(\mathbf q)
  =
  \mathbf p
  =
  (x_{\mathrm{fwd}},y_{\mathrm{lft}})
  =
  (r\cos\theta,r\sin\theta).
\end{equation}

During rollout, the state is updated in Cartesian coordinates. 
To query the sector field at a Cartesian state $\mathbf p=(x_{\mathrm{fwd}},y_{\mathrm{lft}})$, we use the inverse mapping
\begin{equation}
  r
  =
  \sqrt{x_{\mathrm{fwd}}^2+y_{\mathrm{lft}}^2},
  \qquad
  \theta
  =
  \mathrm{atan2}(y_{\mathrm{lft}},x_{\mathrm{fwd}}),
\end{equation}
followed by normalization:
\begin{equation}
  \Pi_{\mathrm{polar}}(\mathbf p)
  =
  \left(
  \tilde{\theta},
  \tilde r
  \right)
  =
  \left(
  \frac{2\theta}{\mathrm{HFoV}},
  \frac{r}{z_{\max}}
  \right).
\end{equation}
The resulting coordinate is clipped to $\Omega$ when necessary.

Importantly, this conversion applies only to query locations. 
The predicted field value is always an ego-BEV Cartesian velocity:
\begin{equation}
  \mathbf v_{\phi}(\mathbf q\mid I,D,\ell)
  =
  (v_{\mathrm{fwd}},v_{\mathrm{lft}}),
\end{equation}
not a polar velocity such as $(\dot{\tilde\theta},\dot{\tilde r})$. 
Therefore, rollout queries the field in normalized polar coordinates but integrates motion in Cartesian coordinates.

\subsection{Vision--Language Encoder (Depth Module Included)}
\label{app:cofls_encoder}

The encoder maps RGB, optional depth, and language into fused action context tokens. 
We use an unfrozen SigLIP 2 backbone $(f_{\mathrm{vis}},f_{\mathrm{txt}})$ for visual and text feature extraction. 
Given the RGB observation $I$ and local command $\ell$, we extract
\begin{equation}
  \mathbf Z_v = f_{\mathrm{vis}}(I)
  \in\mathbb R^{N_v\times d_{\mathrm{emb}}},\quad
  \mathbf Z_\ell = f_{\mathrm{txt}}(\ell)
  \in\mathbb R^{N_\ell\times d_{\mathrm{emb}}},
\end{equation}
where $N_v$ is the number of visual tokens, $N_\ell$ is the number of text tokens, and $d_{\mathrm{emb}}$ is the embedding dimension.

Depth is injected as bounded geometric modulation of the visual tokens. 
The depth map is first normalized to the active sector range and patchified by a convolution whose kernel size and stride match the RGB patch size. 
This produces depth features aligned with the visual tokens:
\begin{equation}
  \mathbf Z_d
  =
  f_{\mathrm{dep}}(D,M_D)
  \in\mathbb R^{N_v\times d_{\mathrm{dep}}},
\end{equation}
where $M_D$ is an optional depth-valid mask. 
When the valid mask is not provided, it is derived from the sector depth range:
\begin{equation}
  M_D = \mathbf{1}\!\left[0<D<z_{\max}\right].
\end{equation}
The valid mask is concatenated with the depth map as an additional input channel before patchification.

After LayerNorm, a small MLP predicts FiLM parameters:
\begin{equation}
  (\gamma,\delta)
  =
  f_{\mathrm{FiLM}}
  \left(
  \mathrm{LN}(\mathbf Z_d)
  \right),
  \qquad
  \gamma,\delta
  \in
  \mathbb R^{N_v\times d_{\mathrm{emb}}}.
\end{equation}
To avoid over-modulating the semantic RGB representation, CoFL-S uses bounded residual FiLM:
\begin{equation}
  \bar{\mathbf Z}_v
  =
  \left(
  1+\iota\tanh(\gamma)
  \right)
  \odot
  \mathbf Z_v
  +
  \eta\tanh(\delta),
  \label{eq:bounded_depth_film}
\end{equation}
where the effective scale and shift gains are
\begin{equation}
  \iota=\iota_{\max}\sigma(a_\phi),
  \qquad
  \eta=\eta_{\max}\sigma(b_\phi),
\end{equation}
where $a_\phi$ and $b_\phi$ are learnable scalar logits, ensuring 
$\iota\in(0,\iota_{\max})$ and $\eta\in(0,\eta_{\max})$. In the default implementation, $\iota_{\max}=0.1$ and $\eta_{\max}=0.05$. 
The final FiLM layer is initialized to zero, so the initial modulation is close to identity:
\begin{equation}
  \bar{\mathbf Z}_v\approx \mathbf Z_v.
\end{equation}
This keeps RGB tokens as the main semantic carrier, while depth reshapes them into geometry-aware visual tokens. 
If depth is not used, we set $\bar{\mathbf Z}_v=\mathbf Z_v$.

The depth-modulated visual tokens and text tokens are projected into a shared hidden dimension $d$:
\begin{equation}
  \tilde{\mathbf Z}_v
  =
  \mathrm{LN}
  \left(
  \mathrm{Projection}_v(\bar{\mathbf Z}_v)
  \right)
  \in\mathbb R^{N_v\times d},\quad
  \tilde{\mathbf Z}_\ell
  =
  \mathrm{LN}
  \left(
  \mathrm{Projection}_\ell(\mathbf Z_\ell)
  \right)
  \in\mathbb R^{N_\ell\times d}.
\end{equation}
The projected streams are fused using $L$ transformer decoder layers. 
Starting from $\mathbf H^{(0)}=\tilde{\mathbf Z}_v$, we apply
\begin{equation}
  \mathbf H^{(l)}
  =
  \mathcal D^{(l)}
  \left(
  \mathbf H^{(l-1)},\tilde{\mathbf Z}_\ell
  \right),
  \qquad
  l\in\{1,\ldots,L\}.
  \label{eq:cofls_vl_decoder}
\end{equation}
Each layer consists of non-causal self-attention over visual tokens, cross-attention with keys and values from $\tilde{\mathbf Z}_\ell$, and an FFN, each wrapped with residual connections and normalization. 
The final context tokens are
\begin{equation}
  \mathbf C
  =
  \mathbf H^{(L)}
  \in\mathbb R^{N_v\times d},
\end{equation}
which provide a language-conditioned representation of the ego-centric scene.

\subsection{CoFL-S Decoder}
\label{app:cofls_decoder}

The decoder adopts a coordinate-query design. 
Given $N$ sector coordinates
\begin{equation}
  \mathbf X
  =
  \{\mathbf q_i\}_{i=1}^{N}
  \in\Omega^{N\times2},
\end{equation}
where the $i$-th row is $\mathbf q_i=(\tilde\theta_i,\tilde r_i)$, we first compute coordinate embeddings and project them to $d$-dimensional query tokens:
\begin{equation}
  \mathbf Q
  =
  \psi(\mathbf X)
  \in\mathbb R^{N\times d}.
  \label{eq:cofls_pos_embed}
\end{equation}
The embedding $\psi$ is implemented as a Gaussian Fourier encoding followed by a linear projection and LayerNorm.

The query tokens attend to the context tokens through $\tilde L$ simplified transformer decoder layers:
\begin{equation}
  \tilde{\mathbf H}^{(\tilde l)}
  =
  \tilde{\mathcal D}^{(\tilde l)}
  \left(
  \tilde{\mathbf H}^{(\tilde l-1)},\mathbf C
  \right),
  \qquad
  \tilde l\in\{1,\ldots,\tilde L\},
  \label{eq:cofls_decoder}
\end{equation}
with $\tilde{\mathbf H}^{(0)}=\mathbf Q$. 
Each decoder layer performs cross-attention from $\tilde{\mathbf H}^{(\tilde l-1)}$ to the context tokens $\mathbf C$, followed by an FFN, with residual connections and normalization. 
Query self-attention is omitted by default to keep dense sector querying efficient.

The decoder predicts a positive magnitude and a unit direction for each query:
\begin{align}
  \mathbf M(\mathbf X)
  &=
  \mathrm{Softplus}
  \left(
  \mathrm{MLP}_{\mathrm{mag}}
  \left(
  \tilde{\mathbf H}^{(\tilde L)}
  \right)
  \right)
  \in\mathbb R_{>0}^{N\times1},
  \label{eq:cofls_mag_head}\\
  \mathbf D(\mathbf X)
  &=
  \frac{
  \mathrm{MLP}_{\mathrm{dir}}
  \left(
  \tilde{\mathbf H}^{(\tilde L)}
  \right)
  }{
  \left\|
  \mathrm{MLP}_{\mathrm{dir}}
  \left(
  \tilde{\mathbf H}^{(\tilde L)}
  \right)
  \right\|_2+\epsilon
  }
  \in\mathbb R^{N\times2}.
  \label{eq:cofls_dir_head}
\end{align}
The final velocity is
\begin{equation}
  \mathbf V(\mathbf X)
  =
  \mathbf M(\mathbf X)\odot\mathbf D(\mathbf X)
  \in\mathbb R^{N\times2},
  \label{eq:cofls_velocity_head}
\end{equation}
where $\odot$ denotes broadcasting element-wise multiplication. 
Each row of $\mathbf V(\mathbf X)$ corresponds to the single-query policy value $\mathbf v_\phi(\mathbf q_i\mid I,D,\ell)$, which is the ego-BEV Cartesian velocity $(v_{\mathrm{fwd}},v_{\mathrm{lft}})$ at the corresponding polar sector query.

\subsection{Action Module}
\label{app:cofls_action_module}

The action module provides a discrete termination signal that complements the continuous sector field. 
It uses the same query-based decoder pattern as Appendix~\ref{app:cofls_decoder}, but replaces spatial coordinate queries with learnable action-ID queries. 
Let
\begin{equation}
  \mathbf A^{(0)}
  =
  [\mathrm a_1,\ldots,\mathrm a_{N_{\mathrm{id}}}]^\top
  \in
  \mathbb R^{N_{\mathrm{id}}\times d}
\end{equation}
denote the learnable action-query table, where each row corresponds to one discrete VLN-CE action ID. 
The action queries attend to the same fused context tokens $\mathbf C$ as the sector decoder:
\begin{equation}
  \mathbf A^{(\hat l)}
  =
  \hat{\mathcal D}^{(\hat l)}
  \left(
  \mathbf A^{(\hat l-1)},\mathbf C
  \right),
  \qquad
  \hat l\in\{1,\ldots,\hat L\}.
  \label{eq:cofls_action_decoder}
\end{equation}
Each layer has the same cross-attention, FFN, residual, and normalization structure as Eq.~\ref{eq:cofls_decoder}. 
The only architectural difference is the query source: $\mathbf Q=\psi(\mathbf X)$ is used for coordinate-conditioned field prediction, whereas $\mathbf A^{(0)}$ is used for action-ID prediction.

After the final action-decoder layer, a row-wise MLP produces one scalar logit for each action query:
\begin{equation}
  \mathbf s
  =
  \mathrm{squeeze}
  \left(
  \mathrm{MLP}_{\mathrm{act}}
  \left(
  \mathbf A^{(\hat L)}
  \right)
  \right)
  \in
  \mathbb R^{N_{\mathrm{id}}}.
  \label{eq:cofls_action_logits}
\end{equation}
The softmax-normalized action probability is
\begin{equation}
  p_{\mathrm{act}}(j)
  =
  \frac{\exp(s_j)}{
  \sum_{m=1}^{N_{\mathrm{id}}}\exp(s_m)
  },
  \qquad
  j\in\{1,\ldots,N_{\mathrm{id}}\}.
  \label{eq:cofls_action_softmax}
\end{equation}
In CoFL-S, this branch is mainly used as a token gate for termination. 

\subsection{Training}
\label{app:cofls_training}

We train CoFL-S by supervising the predicted flow at $N_s$ query locations sampled from replayed VLN-CE episodes. 
For each training episode $e$ and frame $t$, the augmentation pipeline produces
\begin{equation}
  \mathbb D_{e,t} =
  \{(I_{e,t},D_{e,t},\ell_{e,t,k},\mathbf V_{e,t,k}^{*},\tau^{*}_{e,t,k},\mathrm a^{*}_{e,t,k})\}_{k=0}^{K-1},
\end{equation}
Thus, the dataset is organized at the episode level, while the loss is evaluated on sampled frame-slot pairs from each episode.
\subsubsection{Area-Uniform Stratified Sector Sampling}

Dense supervision over all sector cells is expensive. 
We sample $N_s$ query locations with stratified sampling over the sector domain. 
Let a unit square $\mathcal U=[0,1]^2$ be divided into a $g\times g$ grid of cells $\{\mathcal G_j\}_{j=1}^{g^2}$. 
We draw jittered samples
\begin{equation}
  \mathbf u_{j,b}
  =
  (u_{\theta,j,b},u_{r,j,b})
  \sim
  \mathrm U(\mathcal G_j),
  \qquad
  b=1,\ldots,N_{\mathrm{bin}},
  \label{eq:sector_stratified_sampling_u}
\end{equation}
where
\begin{equation}
  N_{\mathrm{bin}}
  =
  \left\lceil
  \frac{N_s}{g^2}
  \right\rceil.
\end{equation}
We concatenate all samples and keep the first $N_s$ points.

A direct uniform sample in $(\tilde r,\tilde\theta)$ would oversample the region near the sector origin, because the polar area element is
\begin{equation}
  dA=r\,dr\,d\theta.
\end{equation}
To obtain approximately area-uniform coverage while preserving stratification, we treat $u_r$ as the radial CDF variable and apply the inverse-CDF transform:
\begin{equation}
  \tilde\theta
  =
  2u_\theta-1,
  \qquad
  \tilde r
  =
  \sqrt{u_r}.
  \label{eq:sector_area_sampling}
\end{equation}
This yields sector queries
\begin{equation}
  \mathbf X
  =
  \{\mathbf q_i\}_{i=1}^{N_s},
  \qquad
  \mathbf q_i=(\tilde\theta_i,\tilde r_i)\in\Omega,
\end{equation}
with balanced coverage over the physical sector area.

\subsubsection{Target Sampling from Cartesian Target Fields}

For sampled sector queries 
\(\mathbf X=\{\mathbf q_i\}_{i=1}^{N_s}\), the decoder predicts velocities at normalized polar coordinates. 
However, the ground-truth target field is stored on an ego-BEV Cartesian raster constructed by the data augmentation pipeline. 
Therefore, the target velocity for query \(\mathbf q_i\) is obtained by sampling the corresponding Cartesian target field at \(\mathbf p_i=\Pi_{\mathrm{cart}}(\mathbf q_i)\):
\begin{equation}
  \mathbf v_i^*
  =
  \mathrm{BilinearSample}
  \left(
  \mathbf V_{e,t,k}^{*},
  \mathbf p_i
  \right)
  =
  (v^*_{\mathrm{fwd}},v^*_{\mathrm{lft}}).
  \label{eq:cart_gt_sampling}
\end{equation}
For brevity, within a fixed frame-slot instance, we write
\begin{equation}
  \mathbf v^*(\mathbf q_i)
  \equiv
  \mathrm{BilinearSample}\left(\mathbf V_{e,t,k}^{*},
  \Pi_{\mathrm{cart}}(\mathbf q_i)
  \right).
\end{equation}
Thus, the model is queried in normalized polar coordinates, while supervision is obtained by mapping those queries to Cartesian positions and sampling Cartesian velocity targets.

\subsubsection{Loss Function}

Given sampled queries $\mathbf X\sim p(\mathbf X)$, the objective combines dense field supervision and action-classification supervision:
\begin{equation}
  \mathcal L
  =
  \mathcal L_{\mathrm{dir}}
  +
  \lambda_\text{mag}
  \mathcal L_{\mathrm{mag}}
  +
  \lambda_\text{act}
  \mathcal L_{\mathrm{act}}.
  \label{eq:cofls_total_loss}
\end{equation}
The direction loss enforces angular alignment via cosine similarity:
\begin{equation}
  \mathcal L_{\mathrm{dir}}
  =
  \mathbb E_{\mathbf X}
  \left[
  \sum_{i=1}^{N_s}
  \left(
  1-
  \frac{
  \mathbf v(\mathbf q_i)^\top
  \mathbf v^*(\mathbf q_i)
  }{
  \|\mathbf v(\mathbf q_i)\|_2
  \|\mathbf v^*(\mathbf q_i)\|_2+\epsilon
  }
  \right)
  \right],
  \label{eq:cofls_dir_loss}
\end{equation}
where $\mathbf v(\mathbf q_i)$ is the prediction corresponding to the $i$-th row of $\mathbf V(\mathbf X)$. 
The magnitude loss matches velocity norms:
\begin{equation}
  \mathcal L_{\mathrm{mag}}
  =
  \mathbb E_{\mathbf X}
  \left[
  \sum_{i=1}^{N_s}
  \left(
  \|\mathbf v(\mathbf q_i)\|_2
  -
  \|\mathbf v^*(\mathbf q_i)\|_2
  \right)^2
  \right].
  \label{eq:cofls_mag_loss}
\end{equation}

The action target is derived from the discrete ground-truth action ID $\mathrm a^*_{e,t,k}$. 
We convert it into a one-hot distribution
\begin{equation}
  p^*_{\mathrm{act}}(j)
  =
  \mathbf{1}
  \left[
  j=\mathrm a^*_{e,t,k}
  \right],
  \qquad
  j\in\{1,\ldots,N_{\mathrm{id}}\}.
  \label{eq:cofls_action_target}
\end{equation}
The action loss is the cross entropy between the predicted softmax probability $p_{\mathrm{act}}$ and the one-hot target $p^*_{\mathrm{act}}$:
\begin{equation}
  \mathcal L_{\mathrm{act}}
  =
  -
  \sum_{j=1}^{N_{\mathrm{id}}}
  p^*_{\mathrm{act}}(j)
  \log
  \left(
  p_{\mathrm{act}}(j)+\epsilon
  \right).
  \label{eq:cofls_action_loss}
\end{equation}
Since $p^*_{\mathrm{act}}$ is one-hot, this is equivalent to the negative log-probability of the annotated action ID. 
The field losses supervise where to move for non-stop execution, while $\mathcal L_{\mathrm{act}}$ teaches the auxiliary branch when a command has an explicit terminal signal.

\subsection{Inference}
\label{app:cofls_inference}

At each closed-loop control cycle, CoFL-S first evaluates the action module on the current observation and command. 
Let $c_{\mathrm{stop}}$ denote the STOP action ID. 
The execution loop terminates before field rollout when the STOP action has the highest probability:
\begin{equation}
  \operatorname*{arg\,max}_{j}
  p_{\mathrm{act}}(j) = c_{\mathrm{stop}}.
  \label{eq:cofls_stop_gate}
\end{equation}
When this condition holds, the current local command is treated as completed and execution terminates. 
Otherwise, CoFL-S generates non-stop motion by numerically integrating the predicted sector field.

For efficiency, the decoder is first queried on a regular polar lattice
\begin{equation}
  \hat{\mathbf X}
  =
  \{\hat{\mathbf q}_{ij}\}_{i=1,j=1}^{\tilde N_\theta,\tilde N_r}
  \subset\Omega
\end{equation}
to obtain a cached velocity grid
\begin{equation}
  \hat{\mathbf V}
  =
  \mathbf V(\hat{\mathbf X})
  \in
  \mathbb R^{\tilde N_\theta\times\tilde N_r\times2}.
\end{equation}
The subsequent rollout uses bilinear interpolation on $\hat{\mathbf V}$.

Given an initial Cartesian position $\mathbf p_0$, we compute the corresponding sector coordinate
\begin{equation}
  \mathbf q_0
  =
  \Pi_{\mathrm{polar}}(\mathbf p_0).
\end{equation}
For a horizon of $T$ steps, we perform forward Euler integration. 
At step $j\in\{0,\ldots,T-1\}$, we first interpolate the velocity at the current sector state:
\begin{equation}
  \mathbf v_j
  =
  \mathrm{BilinearSample}
  \left(
  \hat{\mathbf V},\mathbf q_j
  \right).
\end{equation}
Since the target magnitude is distance-to-go scaled, the raw field velocity decreases as the rollout approaches the local target. 
While this is useful for stable supervision, directly integrating such velocities over a fixed normalized horizon can produce overly short trajectories. 
We therefore apply a bounded inverse-time rescaling during inference:
\begin{equation}
  \tilde{\mathbf v}_j
  =
  s(j,T)\mathbf v_j,
  \qquad
  s(j,T)=
  \frac{1}{
  (1-\frac{j}{T})+\beta (\frac{j}{T})^\alpha
  }.
  \label{eq:cofls_inv_schedule}
\end{equation}
The schedule is motivated by the ideal displacement-field case, where inverse remaining-time scaling converts distance-to-go vectors into fixed-horizon rollout steps. 
The stabilizer $\beta(\cdot)^\alpha$ keeps the factor bounded near the end of rollout, so this rescaling serves as a numerical correction rather than a finite-time convergence guarantee.
We use fixed inference-only schedule parameters $\beta=0.5$ and $\alpha=10$; they are not learned and do not participate in training.

The state is updated in Cartesian coordinates:
\begin{equation}
  \mathbf p_{j+1}
  =
  \mathbf p_j
  +
  \tilde{\mathbf v}_j\Delta t,
  \qquad
  \Delta t=\frac{1}{T}.
  \label{eq:cofls_euler_cart}
\end{equation}
The next query state is obtained by projecting back to the canonical sector:
\begin{equation}
  \mathbf q_{j+1}
  =
  \mathrm{clip}_{\Omega}
  \left(
  \Pi_{\mathrm{polar}}(\mathbf p_{j+1})
  \right).
  \label{eq:cofls_euler_polar}
\end{equation}
The final rollout is the Cartesian trajectory
\begin{equation}
  \tau
  =
  \{\mathbf p_0,\mathbf p_1,\ldots,\mathbf p_T\}.
\end{equation}
For continuous-control deployment, this trajectory is tracked by a low-level controller. 
For VLN-CE compatibility, the first segment or short-horizon rollout can be projected to the closest discrete action in the benchmark action vocabulary.

\section{Continuous-Control Supervision from VLN-CE Replay}
\label{app:data_aug_details}

This appendix details how we convert R2R-CE and RxR-CE training episodes into dense sector flow field supervision.

For each training episode $e$ and replayed frames $t$, the augmentation pipeline produces an episode-level set
\begin{equation}
  \mathbb D_{e,t} =
  \{(I_{e,t},D_{e,t},\ell_{e,t,k},\mathbf V_{e,t,k}^{*},\tau^{*}_{e,t,k},\mathrm a^{*}_{e,t,k})\}_{k=0}^{K-1},
\end{equation}
where $k$ indexes instruction slots.
Here $I_{e,t}$ and $D_{e,t}$ are RGB-D observations, $\ell_{e,t,k}$ is the $k$-th local instruction, $\mathbf V^{*}_{e,t,k}$ is the slot-specific target flow field, $\tau^{*}_{e,t,k}$ is the trajectory target for trajectory-generation baselines, and $\mathrm a^{*}_{e,t,k}$ is the discrete action label for action-token baselines.
All $K$ slots associated with the same frame share the same observation but differ in instruction and target.

\subsection{Visual Observation Generation}
\label{app:visual_observation_generation}

\paragraph{\textbf{VLN-CE official replay.}}
We start from the official R2R-CE and RxR-CE training episodes and replay their official action sequences to obtain time-indexed observations and agent states:
\begin{equation}
  \{(I_{e,t},D_{e,t},S_{e,t},\mathbf T_{e,t},a^{\mathrm{gt}}_{e,t})\}_{t=0}^{T_e}
  =
  \textsc{ReplayOfficialEpisode}(e),
\end{equation}
where $I_{e,t}$ is the ego-centric RGB frame, $D_{e,t}$ is the aligned depth frame, $S_{e,t}$ is the semantic observation used only by the data-generation pipeline, $\mathbf T_{e,t}$ is the agent pose, and $a^{\mathrm{gt}}_{e,t}$ is the official habitat action at frame $t$.
Using the official replay keeps the visual stream, pose sequence, action labels, and fine-grained language annotations aligned with the original VLN-CE trajectory.
Frames with invalid observations, invalid poses, or insufficient local BEV support are discarded.

\subsection{Language Instruction Generation}
\label{app:language_instruction_generation}

\paragraph{\textbf{Fine-grained human anchors.}}
The first instruction slot is a human-written local anchor.
For R2R-CE episodes, we use Fine-Grained R2R \cite{hong2020sub} annotations; for RxR-CE episodes, we use Landmark-RxR \cite{he2021landmark} annotations.
These annotations provide path-aligned sub-instructions associated with local route segments.

For each replayed frame, we localize the agent on the reference path and select the active fine-grained sub-instruction:
\begin{equation}
  \ell_{e,t,0}
  =
  \textsc{SubInstruction}(e,t).
\end{equation}
The anchor goal is not generated by our templates.
Instead, it is taken from the corresponding fine-grained instruction segment:
\begin{equation}
  \mathbf g_{e,t,0}
  =
  \textsc{SubGoal}(e,t).
\end{equation}
In practice, this sub-goal corresponds to the final viewpoint of the active sub-instruction segment, and the associated sub-path is also inherited from the annotation-aligned reference path.
Thus, the anchor slot uses both language and local target information from the fine-grained dataset annotation.
Frames that cannot be assigned to a valid fine-grained sub-instruction are skipped.

\paragraph{\textbf{$K$-slot instruction expansion.}}
For each retained replayed frame, we construct $K$ instruction slots:
\begin{equation}
  \{\ell_{e,t,k}\}_{k=0}^{K-1}.
\end{equation}
The first slot $k=0$ is the fine-grained human anchor.
The remaining slots are sampled from semantically grounded local alternatives:
\begin{equation}
  \{(\ell_{e,t,k},\mathbf g_{e,t,k})\}_{k=1}^{K-1}
  =
  \textsc{SampleAlternatives}
  (I_{e,t},D_{e,t},S_{e,t},\mathbf T_{e,t}).
\end{equation}

\begin{figure}[!t]
\centering
\resizebox{\linewidth}{!}{
\begin{tikzpicture}[
  font=\small,
  >=Latex,
  root/.style={
    draw=black!55,
    fill=rootgray,
    rounded corners=3pt,
    line width=0.8pt,
    align=center,
    minimum width=4.2cm,
    minimum height=0.8cm,
    font=\small\bfseries
  },
  obj/.style={
    draw=objborder,
    fill=objblue,
    rounded corners=3pt,
    line width=0.8pt,
    align=center,
    minimum width=4.4cm,
    minimum height=0.72cm,
    font=\small\bfseries
  },
  reg/.style={
    draw=regborder,
    fill=reggreen,
    rounded corners=3pt,
    line width=0.8pt,
    align=center,
    minimum width=5.2cm,
    minimum height=0.72cm,
    font=\small\bfseries
  },
  ter/.style={
    draw=terborder,
    fill=terorange,
    rounded corners=3pt,
    line width=0.8pt,
    align=center,
    minimum width=3.2cm,
    minimum height=0.72cm,
    font=\small\bfseries
  },
  leaf/.style={
    draw=black!35,
    fill=white,
    rounded corners=3pt,
    line width=0.55pt,
    align=left,
    inner xsep=5pt,
    inner ysep=4pt,
    text width=4.35cm
  },
  regleaf/.style={
    leaf,
    text width=5.15cm
  },
  terleaf/.style={
    leaf,
    text width=3.15cm,
    align=center
  },
  edge/.style={
    draw=black!45,
    line width=0.65pt,
    -{Latex[length=2mm,width=1.4mm]}
  }
]

\node[root] (root) at (0,0) {Instruction Alternatives};

\node[obj] (object) at (-5.6,-1.45) {Object-grounded};
\node[reg] (region) at (0,-1.45) {Region-grounded};
\node[ter] (terminal) at (5.6,-1.45) {Terminal};

\node[leaf, below=0.35cm of object] (toward) {
\textbf{Toward object}\\[-1mm]
{\scriptsize walk/go/head toward \{object\}}\\
{\scriptsize approach \{object\}; stop at/near \{object\}}
};

\node[leaf, below=0.22cm of toward] (past) {
\textbf{Pass object}\\[-1mm]
{\scriptsize go/walk/continue past \{object\}}\\
{\scriptsize pass/walk by \{object\}}
};

\node[regleaf, below=0.35cm of region] (forward) {
\textbf{Forward}\\[-1mm]
{\scriptsize walk/go/move forward; keep going straight}
};

\node[regleaf, below=0.22cm of forward] (turn) {
\textbf{Target-implicit turn}\\[-1mm]
{\scriptsize turn left; turn right}\\[-0.5mm]
{\scriptsize side determined by the reachable target region}
};

\node[regleaf, below=0.22cm of turn] (enter) {
\textbf{Enter region}\\[-1mm]
{\scriptsize enter/go/walk/step into \{region\}}\\
{\scriptsize exit/step/head out into \{region\}}
};

\node[regleaf, below=0.22cm of enter] (turnenter) {
\textbf{Turn and enter}\\[-1mm]
{\scriptsize turn/bear/veer left into \{region\}}\\
{\scriptsize turn/bear/veer right into \{region\}}
};

\node[regleaf, below=0.22cm of turnenter] (exitenter) {
\textbf{Exit--enter}\\[-1mm]
{\scriptsize leave/exit \{cur\_region\} and enter \{region\}}\\
{\scriptsize step/walk/head out of \{cur\_region\}}
};

\node[regleaf, below=0.22cm of exitenter] (turnexit) {
\textbf{Turn and exit--enter}\\[-1mm]
{\scriptsize turn/bear/head left out of \{cur\_region\}}\\
{\scriptsize turn/bear/head right out of \{cur\_region\}}
};

\node[terleaf, below=0.35cm of terminal] (stop) {
\textbf{Stop}\\[-1mm]
{\scriptsize stop here}
};

\draw[edge] (root.south) -- ++(0,-0.45) -| (object.north);
\draw[edge] (root.south) -- (region.north);
\draw[edge] (root.south) -- ++(0,-0.45) -| (terminal.north);

\draw[edge] (object.south) -- (toward.north);
\draw[edge] (toward.south) -- (past.north);

\draw[edge] (region.south) -- (forward.north);
\draw[edge] (forward.south) -- (turn.north);
\draw[edge] (turn.south) -- (enter.north);
\draw[edge] (enter.south) -- (turnenter.north);
\draw[edge] (turnenter.south) -- (exitenter.north);
\draw[edge] (exitenter.south) -- (turnexit.north);

\draw[edge] (terminal.south) -- (stop.north);

\end{tikzpicture}
}
\caption{
Hierarchy of template families used for instruction expansion. 
Object-grounded templates are grounded by visible objects, region-grounded templates by reachable visible regions, and terminal templates by the current agent location.
}
\label{fig:instruction_template_tree}
\end{figure}
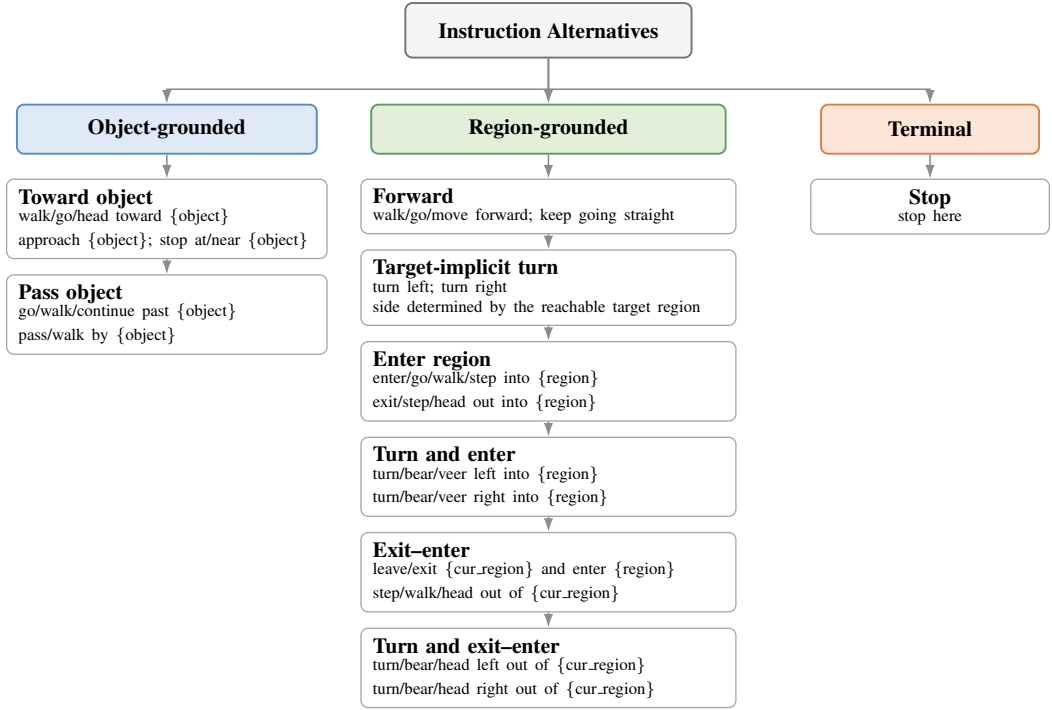

The alternative pool mainly contains object-grounded and region-grounded commands, which are illustrated in Fig.~\ref{fig:instruction_template_tree}.
Object-grounded commands include approaching a visible object and passing a visible object.
Region-grounded commands include entering a visible reachable region, turning before entering a region, or exiting the current region into another visible region.
Optional direction-style commands can be enabled by configuration, while terminal commands are injected only rarely as explicit stop-here alternatives.
Each sampled command is retained only if its goal can be grounded to a valid reachable target and the corresponding target field can be generated.

\subsection{Procedural Annotation}
\label{app:procedural_annotation}

For each replayed frame and instruction slot, we generate an instruction-specific Cartesian BEV target field and the corresponding baseline targets.
The annotation pipeline is summarized in Algorithm~\ref{alg:cofls_gt_generation}.
Given the replayed observations and pose from Appendix~\ref{app:visual_observation_generation}, the pipeline first constructs an ego-centric BEV raster and effective free-space mask, then grounds the instruction to a local goal, and finally derives dense Cartesian velocity supervision through a cost-weighted potential field.

The annotation procedure consists of six stages.

\paragraph{\textbf{Stage 1: Ego-BEV raster and effective free space.}}
For each replayed frame, we construct an ego-centric Cartesian BEV raster centered at the agent. 
The raster coordinate is
\begin{equation}
  \mathbf p=(x_{\mathrm{fwd}},y_{\mathrm{lft}}),
\end{equation}
where $x_{\mathrm{fwd}}$ points forward and $y_{\mathrm{lft}}$ points left.
The raster covers
\begin{equation}
  x_{\mathrm{fwd}}\in[0,x_{\max}],
  \qquad
  y_{\mathrm{lft}}\in[-x_{\max},x_{\max}],
\end{equation}
with resolution $\Delta_{\mathrm{bev}}$.
We rasterize the walkable mask $M_{\mathrm{free}}$ from the navigation geometry and compute the depth-visible mask $M_{\mathrm{vis}}$ from the current observation.
Their intersection defines the effective free-space mask:
\begin{equation}
  M_{\mathrm{eff}}=M_{\mathrm{free}}\wedge M_{\mathrm{vis}}.
\end{equation}
Cells outside $M_{\mathrm{eff}}$ are not removed from the stored field; they are used to construct obstacle and invisible-region escape directions.

\begin{algorithm}[!t]
\caption{Instruction-Specific BEV Annotation Pipeline}
\label{alg:cofls_gt_generation}
\small
\begin{algorithmic}[1]
\Require replayed frame $(I_{e,t},D_{e,t},S_{e,t},\mathbf T_{e,t})$, instruction slot $\ell_{e,t,k}$, slot family $f_{e,t,k}$, official replay action $a^{\mathrm{gt}}_{e,t}$, resolution $\Delta_\mathrm{bev}$
\Ensure target flow field $\mathbf V^{*}_{e,t,k}$, trajectory $\tau^{*}_{e,t,k}$, discrete action $a^{*}_{e,t,k}$
\vspace{0.1cm}
\hrule
\vspace{0.1cm}

\Statex \textbf{\textcolor{blue!70!black}{Stage 1: Ego-BEV Raster and Effective Free Space}}
\State $\mathcal R_{e,t} \gets \textsc{BuildEgoBEVRaster}(\mathbf T_{e,t})$
\Comment{$x_{\mathrm{fwd}}\in[0,x_{\max}]$, $y_{\mathrm{lft}}\in[-x_{\max},x_{\max}]$}
\State $M_{\mathrm{free}} \gets \textsc{RasterizeWalkable}(\mathcal R_{e,t})$
\State $M_{\mathrm{vis}} \gets \textsc{DepthVisible}(D_{e,t},\mathbf T_{e,t})$
\State $M_{\mathrm{eff}} \gets M_{\mathrm{free}}\wedge M_{\mathrm{vis}}$
\State $M_{\mathrm{inv}} \gets \neg M_{\mathrm{eff}}$

\Statex \textbf{\textcolor{blue!70!black}{Stage 2: Instruction-Specific Goal Grounding}}
\If{$k=0$}
   \State $(\ell_{e,t,0},\mathbf g_{e,t,0},\mathcal P_{e,t,0})
   \gets \textsc{FineGrainedAnchor}(e,t)$
   \Comment{Fine-Grained-R2R / Landmark-RxR text, sub-goal, and sub-path}
\Else
   \State $(\ell_{e,t,k},\mathbf g_{e,t,k},f_{e,t,k})
   \gets \textsc{SampleAlternative}(I_{e,t},D_{e,t},S_{e,t},\mathbf T_{e,t})$
   \Comment{object, region, optional direction, or rare terminal}
\EndIf
\State $\mathbf g_{e,t,k}\gets \textsc{SnapReachable}(\mathbf g_{e,t,k},M_{\mathrm{eff}})$

\Statex \textbf{\textcolor{blue!70!black}{Stage 3: Cost-Weighted Geodesic Attraction}}
\State $D_{\mathrm{free}}\gets \textsc{DTO}(M_{\mathrm{eff}})$
\Comment{distance to invalid cells in effective free space}
\State $C_{\mathrm{cost}}(\mathbf p)
\gets 1+\lambda_{\mathrm{safe}}
[\rho_{\mathrm{safe}}-D_{\mathrm{free}}(\mathbf p)]_+$
\State $(D^w_{g,e,t,k},\mathrm{pred}_{e,t,k})
\gets \textsc{Geodesic}(M_{\mathrm{eff}},C_{\mathrm{cost}},\mathbf g_{e,t,k})$
\Comment{cost-weighted distance and predecessor map (Dijkstra~\cite{dijkstra1959note})} 
\State $D^{\mathrm{pix}}_{g,e,t,k}
\gets \textsc{PixelLengthFromPred}(\mathrm{pred}_{e,t,k})$
\State $D^{\mathrm{m}}_{g,e,t,k}
\gets \Delta_{\mathrm{bev}} D^{\mathrm{pix}}_{g,e,t,k}$
\Comment{convert pixel path length to metric distance}

\Statex \textbf{\textcolor{blue!70!black}{Stage 4: Invalid-Region Repulsion and Potential Field}}
\State $D_{\mathrm{obs}}\gets \textsc{DTF}(M_{\mathrm{inv}})$
\Comment{distance to effective free space inside invalid regions}
\State $\Phi_{e,t,k}(\mathbf p)\gets
\begin{cases}
w_gD^w_{g,e,t,k}(\mathbf p), & \mathbf p\in M_{\mathrm{eff}},\\
w_{\mathrm{obs}}D_{\mathrm{obs}}(\mathbf p)+b_{\mathrm{obs}}, & \mathbf p\in M_{\mathrm{inv}}.
\end{cases}$

\Statex \textbf{\textcolor{blue!70!black}{Stage 5: Cartesian BEV Velocity Field}}
\State $\mathbf u_{e,t,k}(\mathbf p)
\gets -\nabla\Phi_{e,t,k}(\mathbf p)/
(\|\nabla\Phi_{e,t,k}(\mathbf p)\|+\epsilon)$
\If{$\mathbf p\in M_{\mathrm{eff}}$}
   \State $\mathbf V^{*}_{e,t,k}(\mathbf p)
   \gets s(D^{\mathrm{m}}_{g,e,t,k}(\mathbf p))\,\mathbf u_{e,t,k}(\mathbf p)$
   \Comment{distance-to-go scaling}
\Else
   \State $\mathbf V^{*}_{e,t,k}(\mathbf p)
   \gets s_{\mathrm{esc}}\,\mathbf u_{e,t,k}(\mathbf p)$
   \Comment{escape velocity}
\EndIf
\Statex \textbf{\textcolor{blue!70!black}{Stage 6: Baseline Target Extraction}}
\State $\tau^{*}_{e,t,k}
\gets \textsc{BacktrackAndResample}(\mathrm{pred}_{e,t,k})$
\Comment{Dijkstra Cartesian trajectory}
\If{$k=0$ and $a^{\mathrm{gt}}_{e,t}$ is available}
   \State $a^{*}_{e,t,0}\gets \textsc{MapHabitatAction}(a^{\mathrm{gt}}_{e,t})$
   \Comment{official replay action}
\Else
   \State $a^{*}_{e,t,k}\gets \textsc{BucketTrajectory}(\tau^{*}_{e,t,k},f_{e,t,k})$
\EndIf

\State \Return $(\mathbf V^{*}_{e,t,k},\,\tau^{*}_{e,t,k},\,a^{*}_{e,t,k})$
\end{algorithmic}
\end{algorithm}

\paragraph{\textbf{Stage 2: Instruction-specific goal grounding.}}
For the anchor slot $k=0$, the instruction, goal, and sub-path are taken from the fine-grained annotation stream. 
For R2R-CE, we use Fine-Grained-R2R; for RxR-CE, we use Landmark-RxR. 
The anchor goal corresponds to the endpoint of the active sub-instruction segment, rather than a synthetic target generated by our templates. 
For $k>0$, we sample semantically grounded alternatives from visible and reachable local scene elements. 
The main families are object-grounded and region-grounded commands, with optional direction-style commands and rare terminal commands. 
Each goal is snapped to a reachable BEV cell before field generation.

\paragraph{\textbf{Stage 3: Cost-weighted geodesic attraction.}}
To encourage paths that stay away from obstacles and invisible boundaries, we construct a traversal cost from the distance-to-invalid transform:
\begin{equation}
  C_{\mathrm{cost}}(\mathbf p)
  =
  1+
  \lambda_{\mathrm{safe}}
  [\rho_{\mathrm{safe}}-D_{\mathrm{free}}(\mathbf p)]_+,
  \label{eq:cofls_costmap}
\end{equation}
where $[z]_+=\max(0,z)$.
Cells closer than the safety radius $\rho_{\mathrm{safe}}$ receive larger traversal costs.
A multi-source Dijkstra search from the instruction-specific goal gives the cost-weighted distance-to-go field $D^w_{g,e,t,k}$ and the predecessor map $\mathrm{pred}_{e,t,k}$.

We also compute the predecessor-path length $D^{\mathrm{pix}}_{g,e,t,k}$ in raster cells and convert it to metric distance by
\begin{equation}
  D^{\mathrm{m}}_{g,e,t,k}(\mathbf p)
  =
  \Delta_{\mathrm{bev}}
  D^{\mathrm{pix}}_{g,e,t,k}(\mathbf p).
  \label{eq:cofls_metric_geodesic}
\end{equation}

\paragraph{\textbf{Stage 4: Invalid-region repulsion and potential field.}}
We compute $D_{\mathrm{obs}}$, the distance from invalid cells to the nearest effective free-space cell.
The composed potential is
\begin{equation}
  \Phi_{e,t,k}(\mathbf p)=
  \begin{cases}
  w_gD^w_{g,e,t,k}(\mathbf p), 
  & \mathbf p\in M_{\mathrm{eff}},\\
  w_{\mathrm{obs}}D_{\mathrm{obs}}(\mathbf p)+b_{\mathrm{obs}},
  & \mathbf p\in M_{\mathrm{inv}}.
  \end{cases}
  \label{eq:cofls_potential}
\end{equation}
The free-space component attracts the agent toward the local goal along cost-weighted geodesic routes, while the invalid-region component provides outward gradients from obstacles, invisible cells, and local boundaries.

\paragraph{\textbf{Stage 5: Cartesian BEV velocity field.}}
The direction field is obtained from the negative potential gradient:
\begin{equation}
  \mathbf u_{e,t,k}(\mathbf p)
  =
  -
  \frac{
  \nabla\Phi_{e,t,k}(\mathbf p)
  }{
  \|\nabla\Phi_{e,t,k}(\mathbf p)\|+\epsilon
  }.
  \label{eq:cofls_flow_dir}
\end{equation}
The final target field is defined directly as
\begin{equation}
  \mathbf V^{*}_{e,t,k}(\mathbf p)
  =
  \begin{cases}
  s(D^{\mathrm{m}}_{g,e,t,k}(\mathbf p))\,
  \mathbf u_{e,t,k}(\mathbf p),
  & \mathbf p\in M_{\mathrm{eff}},\\
  s_{\mathrm{esc}}\,\mathbf u_{e,t,k}(\mathbf p),
  & \mathbf p\in M_{\mathrm{inv}}.
  \end{cases}
  \label{eq:cofls_target_velocity}
\end{equation}
The first branch scales free-space motion by the remaining metric distance-to-go, while the second branch assigns an escape magnitude in invalid regions rather than setting the field to zero.
Since rollout is performed over a normalized time interval, $\mathbf V^{*}$ is a metric displacement rate per unit normalized rollout time, not a direct physical base-velocity command.
The final physical velocity limits are imposed by the shared tracking controller.

\paragraph{\textbf{Stage 6: Baseline target extraction.}}
The trajectory target $\tau^{*}_{e,t,k}$ is obtained by backtracking the Dijkstra predecessor map and resampling the resulting Cartesian path to a fixed horizon.
Thus, trajectory-generation baselines receive a Dijkstra shortest-path target in the ego-centric ground-plane Cartesian frame, not a rollout obtained by integrating the dense field.
For action-token baselines, the Cartesian trajectory is bucketed into a discrete action label using a fixed lookahead radius and a $\pm15^\circ$ forward threshold.
Stop labels for alternatives are emitted only for explicit terminal commands or for object-approach commands whose upstream stop-snap collapses the path length to near zero.

\section{Experiments}

\subsection{Implementation Details for Benchmark Experiments}
\label{app:exa_bench}
\subsubsection{Simulation Protocol}
\label{app:implementation_details}

\begin{wrapfigure}[10]{r}{0.43\linewidth}
  \vspace{-2em}
  \centering
  \includegraphics[width=\linewidth]{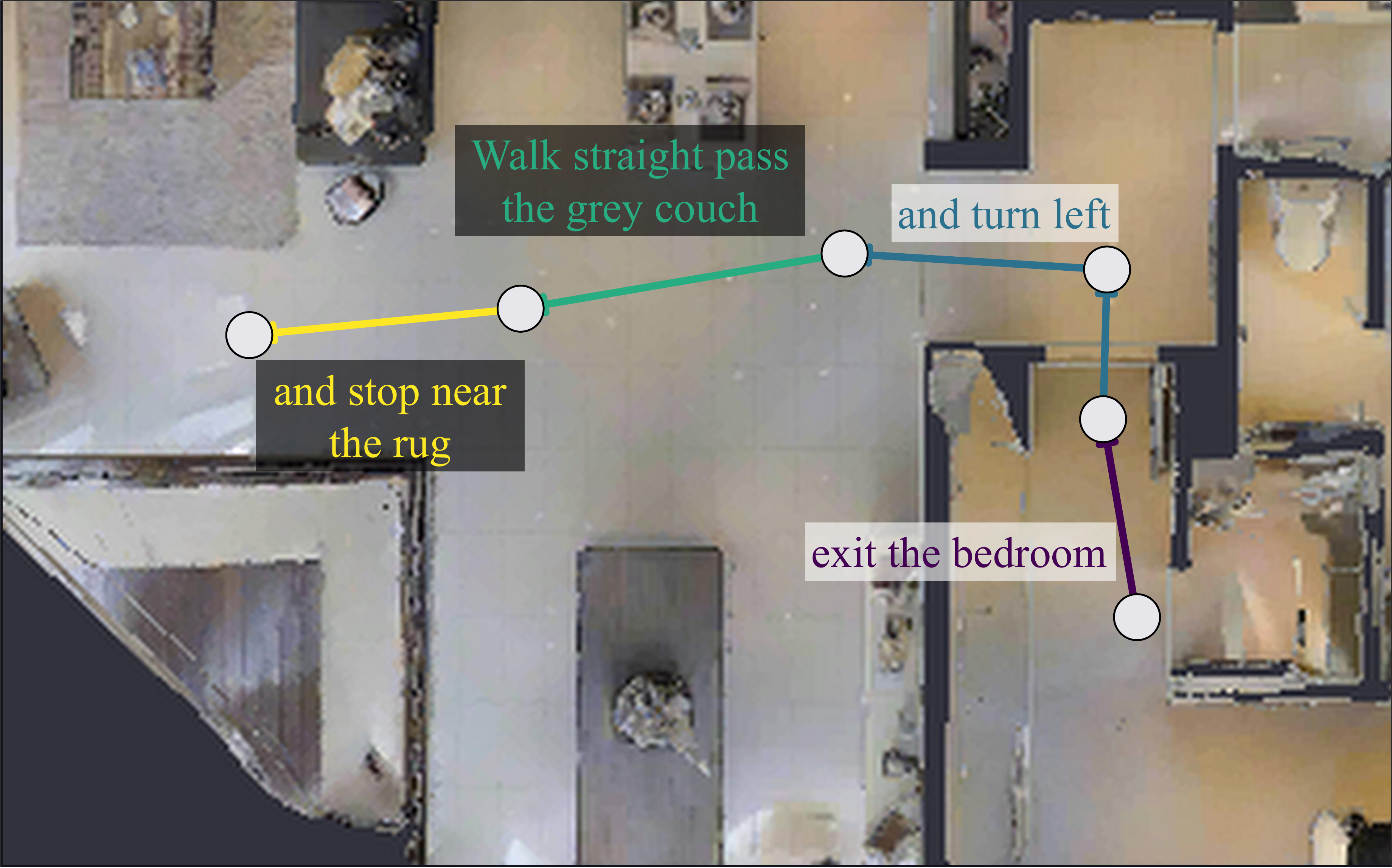}
  \vspace{-1.5em}
  \caption{
    Example of local sub-instruction alignment.
  }
  \label{fig:topdown_aligned}
  \vspace{-1.0em}
\end{wrapfigure}

\paragraph{Supplements for training setup.}
For each episode, we use path-aligned fine-grained annotations to associate the reference route with local language commands.
As illustrated in Fig.~\ref{fig:topdown_aligned}, a long navigation instruction is decomposed into several route segments, each associated with a local sub-instruction.
During benchmark construction, the oracle agent pose is used to identify the current route segment and select the active local sub-instruction.
The policy then receives only the current RGB-D observation and the selected local sub-instruction as input.

All models are trained with matched training settings, including the same number of gradient steps, batch size, optimizer, learning-rate schedule, and input preprocessing. 
RGB-D observations are resized to $224\times224$. 
Depth values are clipped to the range of \qtyrange[range-phrase = --, range-units = single]{0}{5}{\metre}, and pixels outside this range are marked as invalid in the depth mask.

\paragraph{Continuous-time evaluation.}
Although the training data is derived from VLN-CE episodes, evaluation is not performed under the standard discrete VLN-CE action protocol. 
Instead, all methods are evaluated in a continuous-time Habitat environment.
Each policy is queried at a fixed planner frequency (\qty{2},\qty{5},\qty{10}{\hertz}), and the latest velocity command is updated for trajectory tracking until a new command is produced. 
This zero-order-hold execution avoids artificially synchronizing all methods to the slowest policy and allows the measured inference frequency to reflect the actual computational cost of each interface.

The output of each policy is converted into a planar velocity command $(v_t,\omega_t)$ and executed by the same low-level velocity controller. 
The controller applies the same velocity limits, collision handling, and episode termination rules for all methods. 
In continuous-time Habitat evaluation, the simulator updates the agent state at \qty{60}{\hertz} and renders RGB-D observations at \qty{30}{\hertz}. 
We use a maximum linear velocity of \qty{0.5}{\metre\per\second} and a maximum angular velocity of \qty{1.0}{\radian\per\second}. 
Each episode is terminated after \qty{120}{\second} or when the policy predicts STOP. 
We take the standard VLN evaluation protocol as reference; success is defined by a final distance within \qty{3}{\metre} of the goal, and path length is computed by accumulating planar displacement at the \qty{60}{\hertz} simulator rate. 

\subsubsection{Baseline Implementation Details}
\label{app:baseline_details}

\paragraph{Action Token.}
The Action Token baseline follows the standard VLN-CE discrete action interface, predicting four actions: \textsc{Forward}, \textsc{Turn-Left}, \textsc{Turn-Right}, and \textsc{Stop}. 
For architectural matching, this baseline uses the same query-based action module as CoFL-S. 
Unlike CoFL-S, however, the predicted action distribution is used for all low-level decisions rather than only for termination.

At inference time, each non-stop action is realized through the same pure-pursuit tracking controller used by CoFL-S and Action Chunk: 
the planner emits a synthetic two-point local trajectory whose geometry forces the controller into the desired velocity regime,
\begin{equation}
\hat{\tau}_t
=
\begin{cases}
\{(0,0),(L,0)\}, & a_t=\textsc{Forward},\\
\{(0,0),(0,+L)\}, & a_t=\textsc{Turn-Left},\\
\{(0,0),(0,-L)\}, & a_t=\textsc{Turn-Right},
\end{cases}
\end{equation}
where the second waypoint is expressed in the ego-centric ground-plane frame ($x$ forward, $y$ left) and $L>0$ is a fixed synthetic lookahead. 
A \textsc{Forward} waypoint places the lookahead anchor along the forward axis and elicits the controller's nominal forward speed; 
\textsc{Turn-Left} and \textsc{Turn-Right} waypoints place the anchor on the lateral axis, triggering the controller's rotate-in-place branch (see the trajectory tracking controller paragraph) and yielding angular commands of $\pm\omega_{\max}$ with zero linear velocity. 
The synthetic distance $L$ does not affect the command as long as the synthetic waypoint is selected as the lookahead point, because the controller branch depends on the heading error $e_t$ rather than the waypoint distance.

The episode terminates when the predicted \textsc{Stop} probability is the highest among the four actions or when the maximum episode length is reached.

\paragraph{Action Chunk.}
The Action Chunk baseline predicts a fixed-horizon sequence of local displacement vectors from the encoded RGB-D observation and local sub-instruction:
\begin{equation}
\Delta\hat{\tau}_t^{\mathrm{chunk}}
=
\{\Delta\hat{\mathbf p}_{t,1},\ldots,\Delta\hat{\mathbf p}_{t,H}\},
\qquad
\Delta\hat{\mathbf p}_{t,h}\in\mathbb R^2,
\end{equation}
where each displacement is represented in the ego-centric ground-plane frame ($x$ forward, $y$ left). 
These displacement vectors can be interpreted as velocity-like trajectory increments and are accumulated to obtain a local trajectory:
\begin{equation}
\hat{\mathbf p}_{t,h}
=
\sum_{j=1}^{h}\Delta\hat{\mathbf p}_{t,j},
\qquad
\hat{\tau}_t^{\mathrm{chunk}}
=
\{\hat{\mathbf p}_{t,1},\ldots,\hat{\mathbf p}_{t,H}\}.
\end{equation}

To avoid disadvantaging the trajectory-generation baseline with a weaker conditioning architecture, we implement Action Chunk with a Transformer-style decoder rather than a U-Net-style global conditioning module. 
Specifically, the chunk queries attend directly to the full sequence of multimodal encoder tokens, following the same token-level conditioning interface used by CoFL-S. 
Thus, Action Chunk preserves access to fine-grained visual-language tokens, while differing from CoFL-S mainly in the action representation: it predicts a finite current-state-anchored sequence of trajectory increments, whereas CoFL-S predicts a spatially queryable sector field that can be rolled out from arbitrary query states.

At inference time, the accumulated trajectory $\hat{\tau}_t^{\mathrm{chunk}}$ is passed to the shared tracking controller. 
Termination is determined by the same STOP head used in CoFL-S.

\paragraph{Waypoint + GT Planner.} The Waypoint + GT Planner baseline adopts a waypoint-based navigation interface, combining learned endpoint prediction with privileged geometric planning. Given the encoded RGB-D observation and local sub-instruction, it predicts a single future trajectory endpoint $\hat{\mathbf g}_t\in\mathbb R^2$ in the ego-centric ground-plane frame ($x$ forward, $y$ left). The supervision target is the endpoint of the same horizon-$H$ trajectory used by Action Chunk.
Thus, the predicted waypoint represents a finite-horizon motion target rather than necessarily the final goal of the instruction.

For architectural matching, this baseline uses the same Transformer-style diffusion-head architecture as Action Chunk, with the denoising target reduced from a sequence of displacement vectors to a single endpoint.
The endpoint query attends directly to the full sequence of multimodal encoder tokens, preserving the same token-level conditioning interface.

At inference time, the predicted endpoint is transformed into the simulator's world frame, and Habitat's planner generates a path to it using the ground-truth navigation mesh.
The resulting path is transformed back into the ego-centric frame for local trajectory tracking.
This baseline therefore evaluates learned waypoint prediction with access to privileged environment geometry during planning; the waypoint itself remains a model prediction.

\paragraph{Velocity Regression.} The Velocity Regression baseline adopts a direct continuous velocity prediction interface. Given the encoded RGB-D observation and local sub-instruction, it predicts a single instantaneous motion vector: $\hat{\mathbf v}_t=\bigl(\hat v_{t,\mathrm{fwd}},\hat v_{t,\mathrm{lft}}\bigr)\in\mathbb R^2,$ where the two components represent forward and lateral translational velocities in the ego-centric ground-plane frame.
Neither component represents angular velocity.

For architectural matching, the prediction head follows the query-based design of the CoFL-S decoder.
Two learnable queries cross-attend to the full sequence of encoder tokens, and their resulting features are passed through MLP readouts to predict the respective velocity components.
Unlike Action Chunk, this baseline does not predict a future motion sequence; unlike CoFL-S, it does not condition its output on spatial query coordinates.

At inference time, the predicted velocity direction defines a lookahead point at a fixed distance from the robot, which is passed to the shared tracking controller to generate linear and angular velocity commands.

\paragraph{Trajectory tracking controller.}
For all methods, we use the same low-level tracking rule based on geometric pure pursuit.
Given a predicted local trajectory $\hat{\tau}_t$, the controller selects a lookahead point $\mathbf p^{\star}=(p^{\star}_x,p^{\star}_y)$ as the first waypoint lying ahead of the agent at planar distance $\geq L_\mathrm{pp}$, falling back to the last forward waypoint when no such point exists.
The heading error in the agent's local ground-plane frame is
\begin{equation}
e_t = \mathrm{atan2}(p^{\star}_y,p^{\star}_x).
\end{equation}
A non-holonomic ground robot cannot recover from a backward lookahead under the geometric law alone, so the controller switches to a rotate-in-place branch when the heading error is large:
\begin{equation}
(v_t,\omega_t)=
\begin{cases}
\bigl(\,0,\ \mathrm{sign}(e_t)\cdot\omega_{\max}\,\bigr),
& |e_t|>e_{\mathrm{rot}},\\[2pt]
\bigl(\,v_{\max}\cdot\max(\cos e_t,0),\ \mathrm{clip}(v_t\,\kappa_t,\,\pm\omega_{\max})\,\bigr),
& \text{otherwise,}
\end{cases}
\end{equation}
where the steering curvature follows the standard pure-pursuit form
\begin{equation}
\kappa_t = \frac{2\sin e_t}{L_\mathrm{pp}}.
\end{equation}
The cosine modulation of $v_t$ slows the agent down when the lookahead drifts off-axis, while the curvature term induces the corresponding yaw rate; both saturate at $v_{\max}$ and $\omega_{\max}$. 
The same values of $L_\mathrm{pp}$, $v_{\max}$, $\omega_{\max}$, and $e_{\mathrm{rot}}$ are used for all methods. 
In all reported experiments we set $L_\mathrm{pp}=$ \qty{0.5}{\metre}, $v_{\max}=$ \qty{0.5}{\metre\per\second}, $\omega_{\max}=$ \qty{1.0}{\radian\per\second}, and $e_{\mathrm{rot}}=$ \qty{0.5}{\radian}.

\subsubsection{Metric Definitions}
\label{app:metric_definitions}

\paragraph{Standard VLN metrics.}
We report standard VLN goal-reaching metrics, including Navigation Error (NE), Oracle Success (OS), Success Rate (SR), and Success weighted by Path Length (SPL), following the VLN-CE convention.

\textbf{Blocked-Step Rate} (BSR) measures how often the agent is blocked by the navigation mesh during execution.
Let $b_n$ denote the number of blocked simulator substeps at control step $n$.
A control step is counted as blocked if $b_n>0$.
BSR is defined as
\begin{equation}
\mathrm{BSR}=\frac{1}{N_{\mathrm{ctrl}}}\sum_{n=1}^{N_{\mathrm{ctrl}}}\mathbf 1[b_n>0],
\end{equation}
where $N_\mathrm{ctrl}$ is the total number of control steps in the episode.
Lower BSR indicates fewer blocked interactions with the navigation mesh.

\textbf{Heading Smoothness} (HS) measures the smoothness of the executed path based on heading changes.
Given the executed trajectory projected onto the ground plane, we first compute the displacement vector between consecutive positions:
\begin{equation}
\mathbf d_t = \mathbf x_{t+1}^{xz} - \mathbf x_t^{xz}.
\end{equation}
Segments with negligible displacement are ignored.
For each valid segment, we compute its heading angle:
\begin{equation}
\psi_t = \mathrm{atan2}(d_{t,z}, d_{t,x}).
\end{equation}
The heading difference between adjacent valid segments is wrapped to $[-\pi,\pi]$.
HS is then computed as
\begin{equation}
\mathrm{HS}
=
\max\left(
0,\,
1-
\frac{1}{\pi}
\cdot
\frac{1}{T_{\psi}}
\sum_{t}
|\Delta \psi_t|
\right),
\end{equation}
where $T_{\psi}$ is the number of valid heading differences.
A straight trajectory has $\mathrm{HS}=1$, while trajectories with frequent sharp heading changes receive lower scores.
\clearpage
\subsubsection{Additional Qualitative Snapshots}
\label{app:more_benchmark_example}
\begin{figure}[h]
  \includegraphics[width=\linewidth]{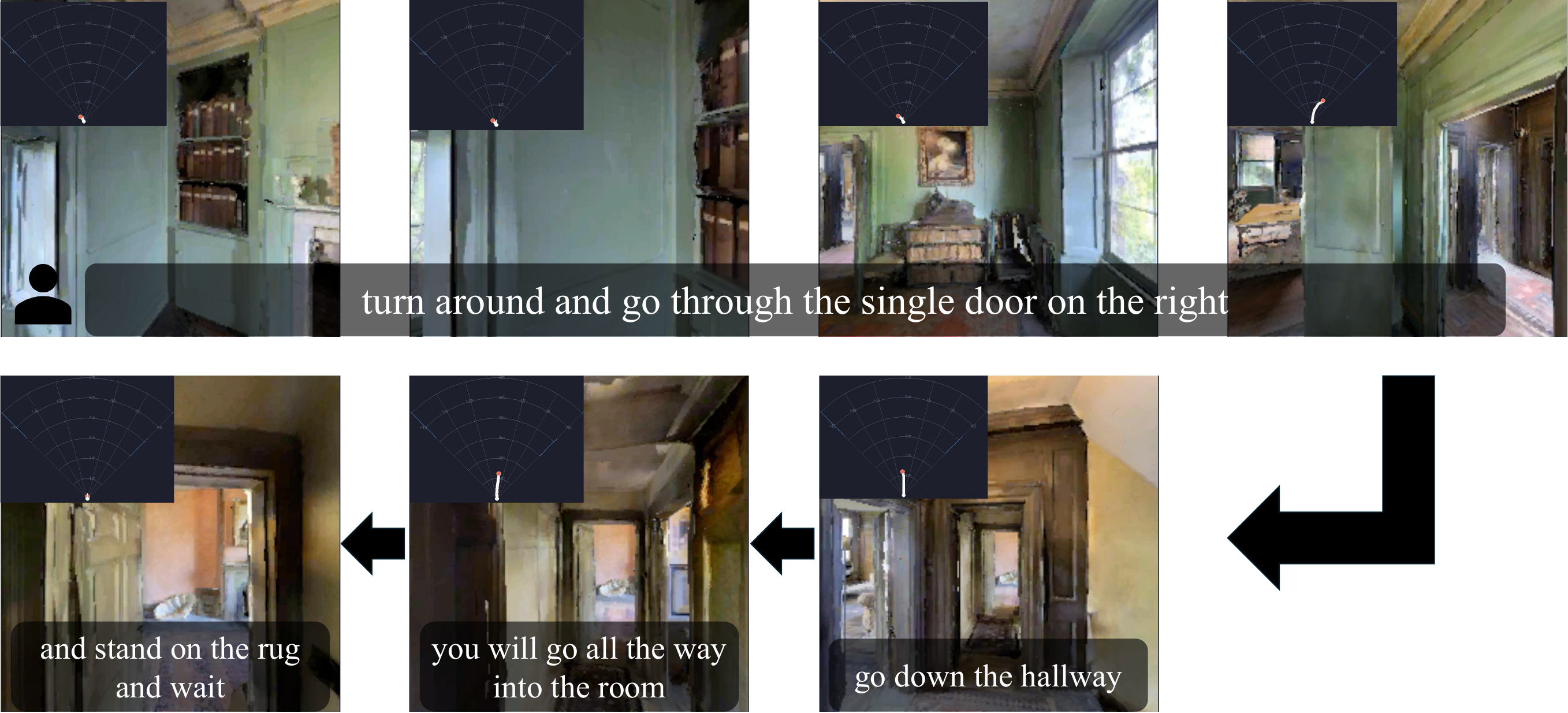}
  \vspace{-1.5em}
  \caption{
    Qualitative snapshots of CoFL-S executing a task sequence in R2R-CE, composed of fine-grained sub-instructions in a Habitat environment~\cite{savva2019habitat}.
    For each step, we visualize the RGB observation, the aligned local instruction, and the predicted trajectory integrated from the sector flow field.
  }
  \label{fig:benchmark_ep29}
\end{figure}

\begin{figure}[h]
  \includegraphics[width=\linewidth]{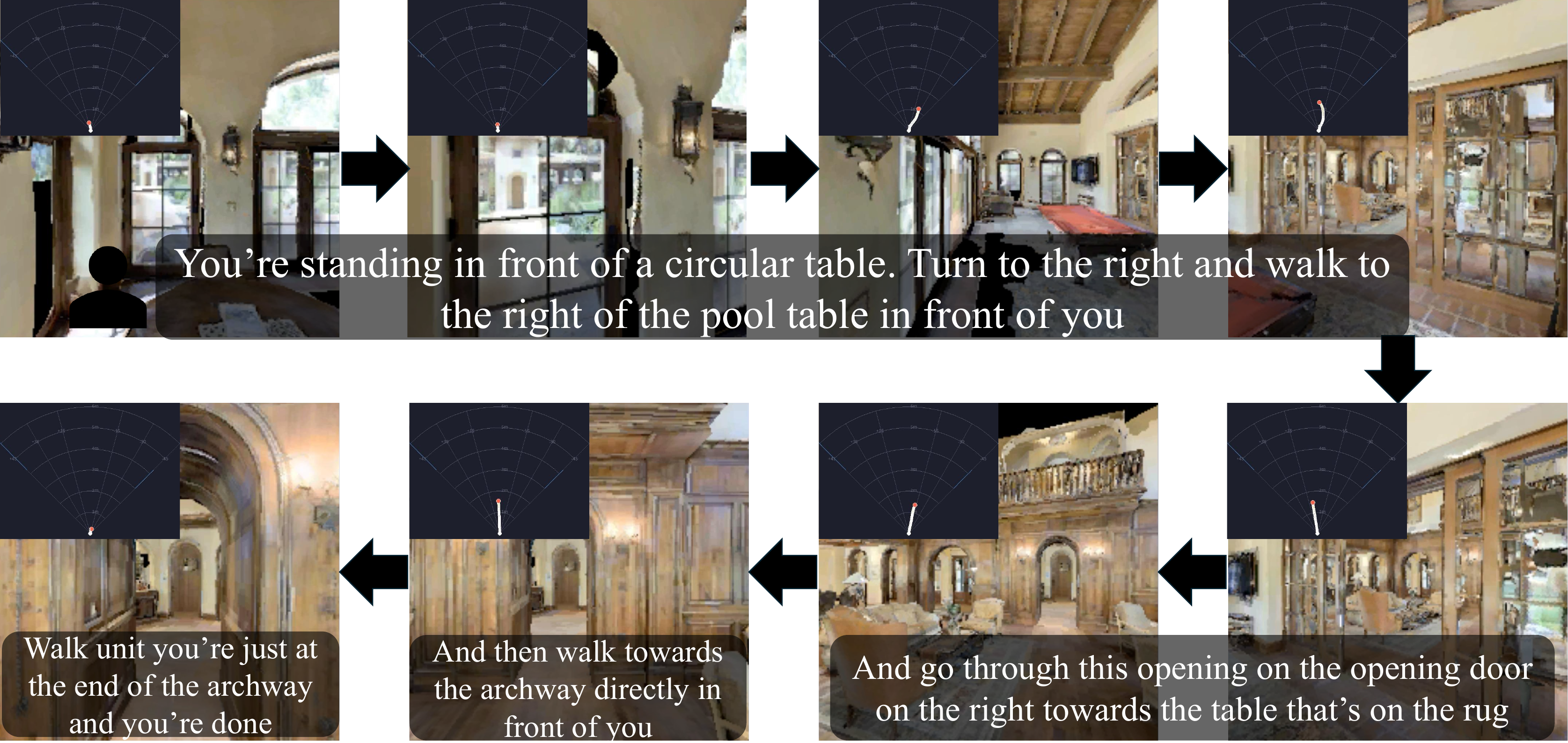}
  \vspace{-1.5em}
  \caption{
    Qualitative snapshots of CoFL-S executing a task sequence in RxR-CE, composed of fine-grained sub-instructions in a Habitat environment~\cite{savva2019habitat}.
    For each step, we visualize the RGB observation, the aligned local instruction, and the predicted trajectory integrated from the sector flow field.
  }
  \label{fig:benchmark_ep6003}
\end{figure}

More qualitative snapshots of CoFL-S are provided in Fig.~\ref{fig:benchmark_ep29} and Fig.~\ref{fig:benchmark_ep6003}.

\subsection{Ablation Studies}
\label{app:abla}

\subsubsection{Spatial Supervision Coverage Ablation}
\label{app:coverage_abla}

CoFL~\cite{liu2026cofl} previously demonstrated the benefit of supervising flow fields over the workspace rather than only along demonstrated trajectories.
Here, we revisit this ablation to verify whether the same finding holds for CoFL-S under ego-centric observations and a sector-field representation.

\begin{wrapfigure}[8]{l}{0.65\linewidth}
\centering
\includegraphics[width=0.31\linewidth]{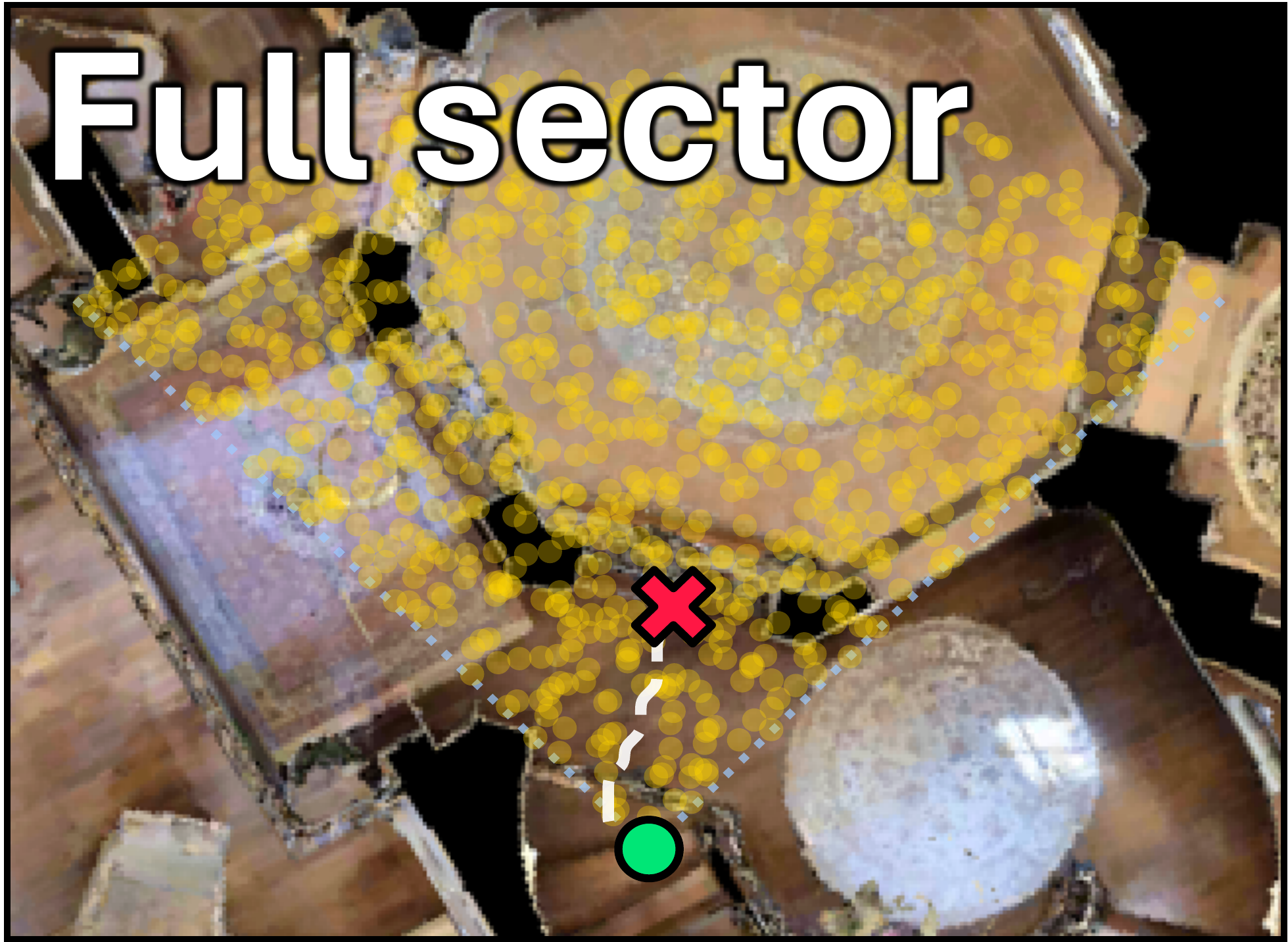}
\hfill
\includegraphics[width=0.31\linewidth]{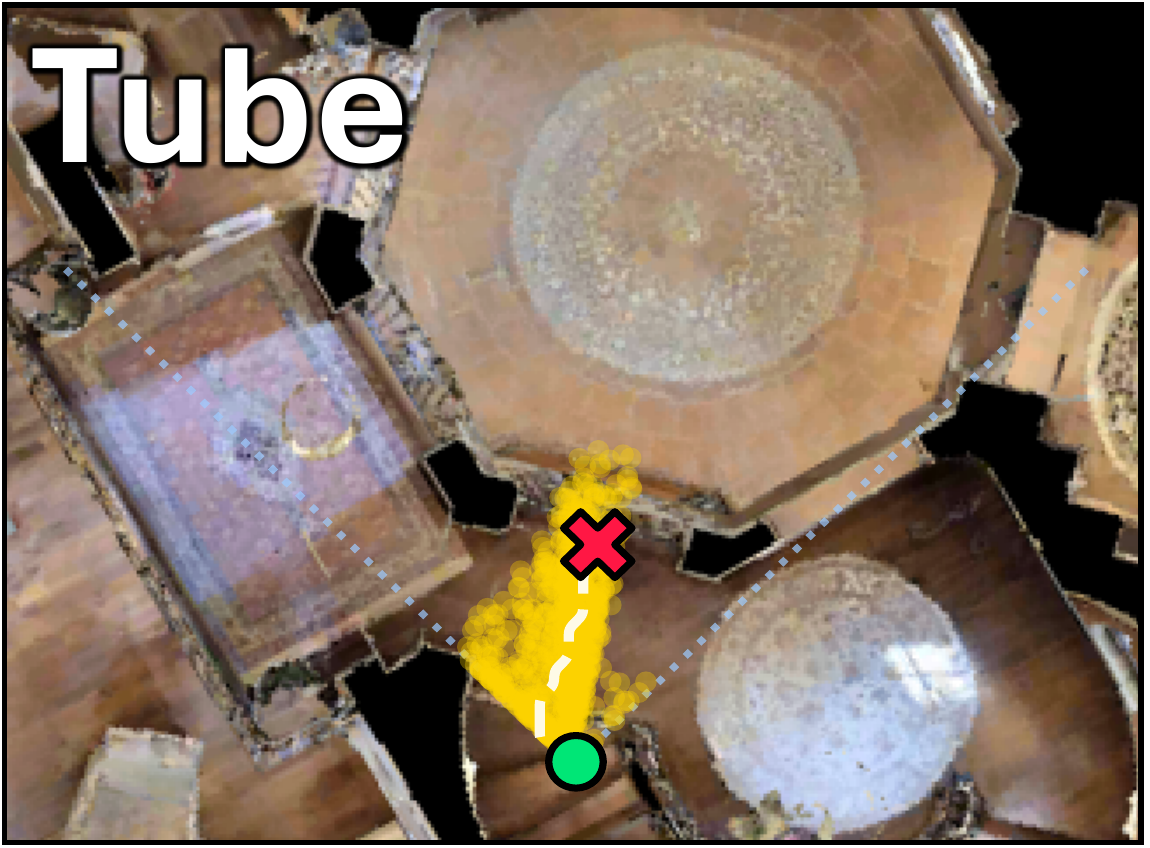}
\hfill
\includegraphics[width=0.31\linewidth]{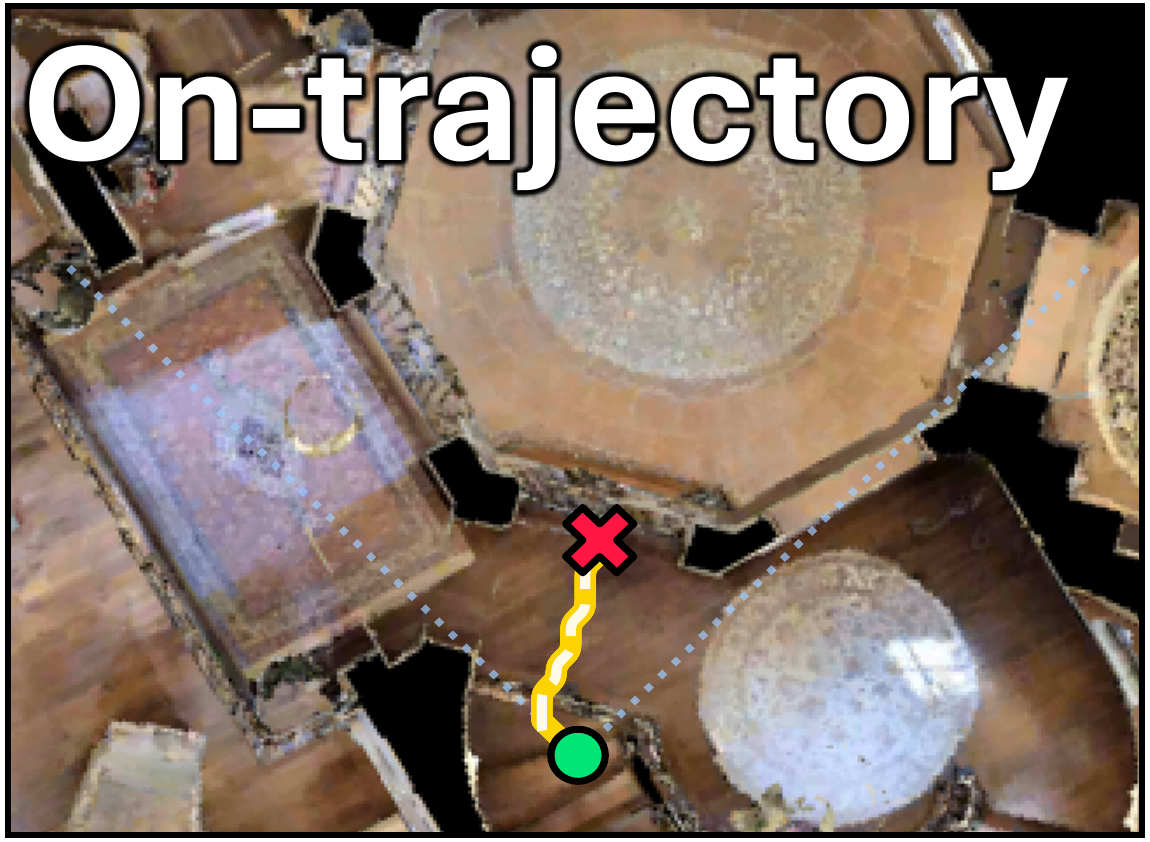}
\caption{
Supervision supports: full sector (left), tube (middle), and on-trajectory (right), with a fixed query budget.
}
\label{fig:supervision_coverage}
\vspace{-0.8em}
\end{wrapfigure}

We compare three supervision supports:
\textbf{On-trajectory} restricts field supervision to the reference trajectory;
\textbf{Tube} extends supervision to a neighborhood around that trajectory;
and \textbf{Full sector} uses the complete sector supervision domain
(Fig.~\ref{fig:supervision_coverage}).
Across variants, we keep the architecture, training samples, query budget, and rollout procedure fixed, varying only the spatial support of field supervision.
This comparison isolates the effect of supervision coverage within the same sector-field representation.

Table~\ref{tab:supervision_coverage} shows that broader supervision coverage consistently improves NE, OS, SR, SPL, and BSR.
Extending supervision from the trajectory to its neighborhood reduces BSR from $0.46$ to $0.19$, but yields only modest gains in task completion.

\begin{wraptable}[9]{r}{0.55\linewidth}
\centering
\setlength{\tabcolsep}{4pt}
\caption{Spatial supervision coverage ablation for CoFL-S.}
\vspace{-0.6em}
\label{tab:supervision_coverage}
\resizebox{\linewidth}{!}{
    \begin{tabular}{lcccccc}
    \toprule
    Support & NE$\downarrow$ & OS$\uparrow$ & SR$\uparrow$ & SPL$\uparrow$ & BSR$\downarrow$ & HS$\uparrow$ \\
    \midrule
    On-trajectory & 10.95 & 0.17 & 0.11 & 0.06 & 0.46 & \underline{0.95} \\
    Tube & \underline{10.26} & \underline{0.23} & \underline{0.13} & \underline{0.12} & \underline{0.19} & 0.92 \\
    \rowcolor{gray!15}
    Full sector & \textbf{7.05} & \textbf{0.49} & \textbf{0.35} & \textbf{0.27} & \textbf{0.15} & \textbf{0.96} \\
    \bottomrule
    \end{tabular}
}
\vspace{-0.5em}
\end{wraptable}

Full-sector supervision further increases SR from $0.13$ to $0.35$ and SPL from $0.12$ to $0.27$.
HS remains within $0.92$--$0.96$ and does not follow the same monotonic trend.
These results corroborate the earlier finding in CoFL~\cite{liu2026cofl}: workspace-level supervision benefits field-based navigation beyond supervision confined to a single trajectory.

The present experiment verifies that this benefit also holds for the ego-centric sector formulation.

\subsubsection{Instruction-diverse Data Ablation}
\label{app:data_abla}
\begin{wraptable}[4]{r}{0.60\linewidth}
\centering
\vspace{-1.5em}
\setlength{\tabcolsep}{4pt}
\caption{
Instruction-diverse data ablation for CoFL-S. 
}
\vspace{-0.6em}
\label{tab:data_ablation}
\resizebox{\linewidth}{!}{
\begin{tabular}{lccccc}
\toprule
Training data & Samples & NE$\downarrow$ & OS$\uparrow$ & SR$\uparrow$ & SPL$\uparrow$ \\
\midrule
Anchor only ($k=0$)        & 2,400,923 & {7.09} & {0.48} & {0.33} & {0.26} \\
\rowcolor{gray!15}
Full augmentation              & 3,758,511 & \textbf{7.05} & \textbf{0.49} & \textbf{0.35} & \textbf{0.27}\\
\bottomrule
\end{tabular}}
\vspace{-0.5em}
\end{wraptable}

We evaluate whether the proposed instruction-diverse training is beneficial for learning language-conditioned sector fields.

We compare two training sets on CoFL-S. 
\textbf{Anchor only} uses only the human sub-instruction slot $k=0$ from Fine-Grained R2R~\cite{hong2020sub} or Landmark-RxR~\cite{he2021landmark}. 
\textbf{Full augmentation} uses the complete augmented dataset. The inference frequency is fixed at \qty{5}{\hertz}.
In Table~\ref{tab:data_ablation}, we report the episode-weighted average results of R2R-CE and RxR-CE. The results show a moderate but consistent gain from instruction-diverse training: the added commands provide useful local semantic variation, but are simpler and not fully distribution-matched to the human-aligned sub-instructions used in simulation evaluation.

\subsubsection{Action and Depth Modulation Ablation}
\label{app:module_abla}

We ablate two optional modules in CoFL-S: the action-token termination module and the depth modulation pathway. 
The action module provides a discrete STOP decision, while all non-stop motions are still generated by integrating the predicted sector field. 
When the action module is removed, termination is approximated by a rollout-stationarity rule: the local command is treated as STOP if the predicted forward displacement is below \qty{0.05}{\metre} and the predicted heading change is below $10^{\circ}$. 
The depth module injects geometric cues into the visual representation. 
\begin{wraptable}[6]{r}{0.45\linewidth}
\centering
\vspace{-0.5em}
\setlength{\tabcolsep}{4pt}
\caption{
Action and depth module ablation for CoFL-S. 
}
\vspace{-0.6em}
\label{tab:module_ablation}
\resizebox{\linewidth}{!}{
\begin{tabular}{cc|ccccc}
\toprule
Action & Depth & NE$\downarrow$ & OS$\uparrow$ & SR$\uparrow$ & SPL$\uparrow$ & BSR$\downarrow$  \\
\midrule
$\checkmark$ &       & 7.41 & 0.46 & 0.30 & 0.23 & 0.27\\
       & $\checkmark$ & 8.13 & 0.28 & 0.19 & 0.15 & \textbf{0.09}\\
\rowcolor{gray!15}
$\checkmark$ & $\checkmark$ & \textbf{7.05} & \textbf{0.49} & \textbf{0.35} & \textbf{0.27} & {0.15} \\
\bottomrule
\end{tabular}}
\vspace{-0.5em}
\end{wraptable}
Table~\ref{tab:module_ablation} compares the three configurations under the same training and evaluation setting, with inference frequency fixed at \qty{5}{\hertz}.

The action module has a large effect on task completion. 
Without it, the heuristic STOP rule often fires before the rollout reaches the local target, especially when the field produces short obstacle-adjacent trajectories; this lowers SR and SPL substantially. 
The low BSR of the depth-only variant should therefore not be interpreted as better safety: many potentially collision-prone rollouts terminate early instead of continuing through the narrow or obstacle-proximal region. 
With reliable termination from the action module, depth modulation reduces BSR from $0.27$ to $0.15$ while also improving NE, OS, SR, and SPL. 
Using both modules gives the strongest overall performance, indicating that discrete stopping and geometric depth cues play complementary roles in the continuous sector-field controller.

\subsection{Supplements for Real-World Experiments}
\label{app:rw}

\subsubsection{Hardware Setup}
\label{app:rw_hardware}

\begin{wrapfigure}[9]{r}{0.45\linewidth}
  \vspace{-5.0em}
  \centering
  \includegraphics[width=\linewidth]{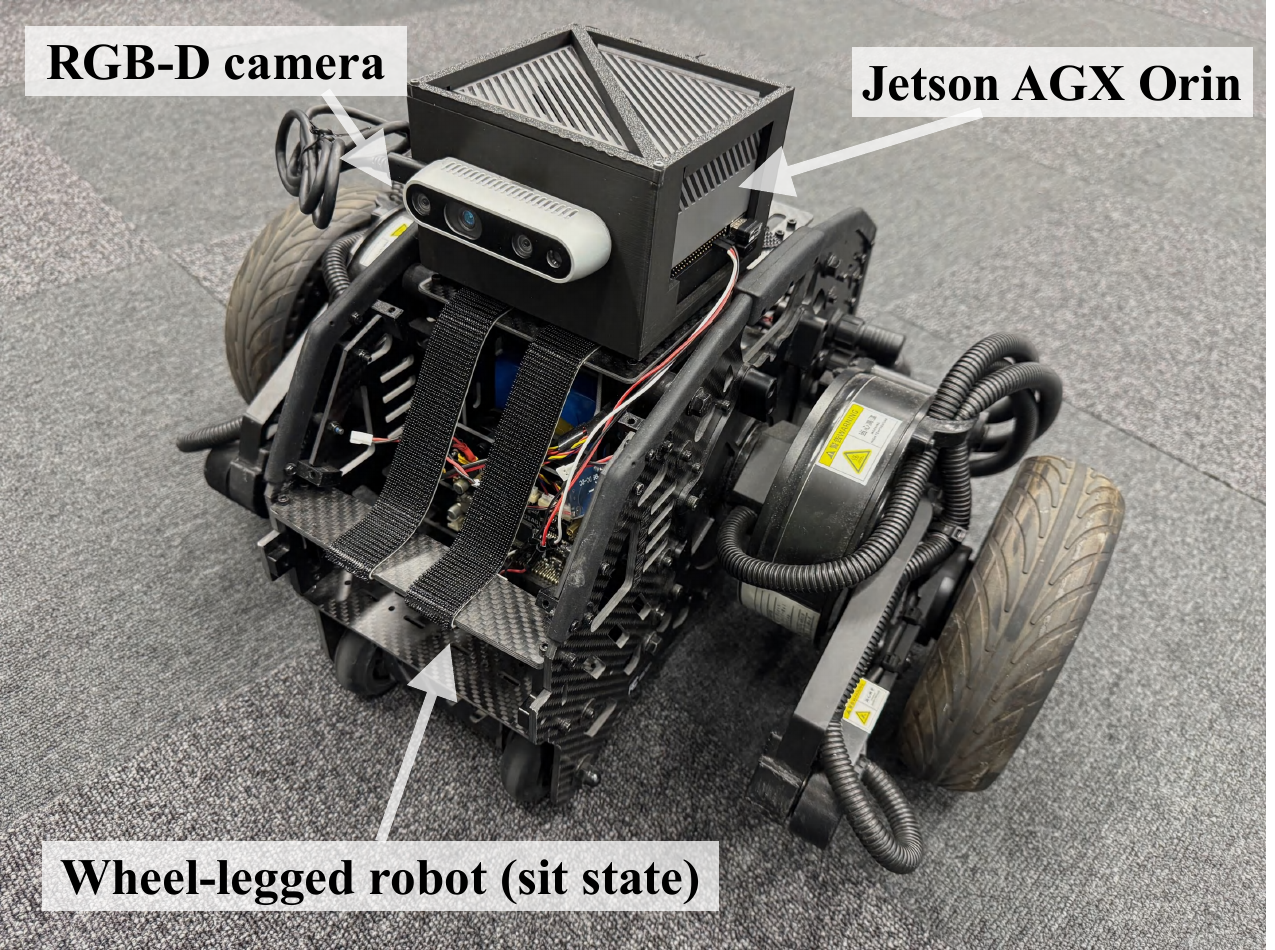}
  \vspace{-1.5em}
  \caption{
    Overview of the hardware setup.
  }
  \label{fig:hardware_setup}
  \vspace{-1.0em}
\end{wrapfigure}

The real-world experiments are conducted on our ground wheel-legged robot platform equipped with an RGB-D camera and an edge computer. 
An illustration of the hardware setup, including the robot platform, camera mounting, and onboard computing unit, is shown in Fig.~\ref{fig:hardware_setup}. Both the CoFL-S model and the trajectory tracker run onboard on the Jetson edge computer. 
At each control step, the tracker converts the model output into a planar velocity command, which is then transmitted to the robot's low-level controller through UART for execution.

\subsubsection{Additional Qualitative Snapshots in Real-World Experiments}
\label{app:more_rw_example}

\begin{figure}[h]
  \centering
  \includegraphics[width=\linewidth]{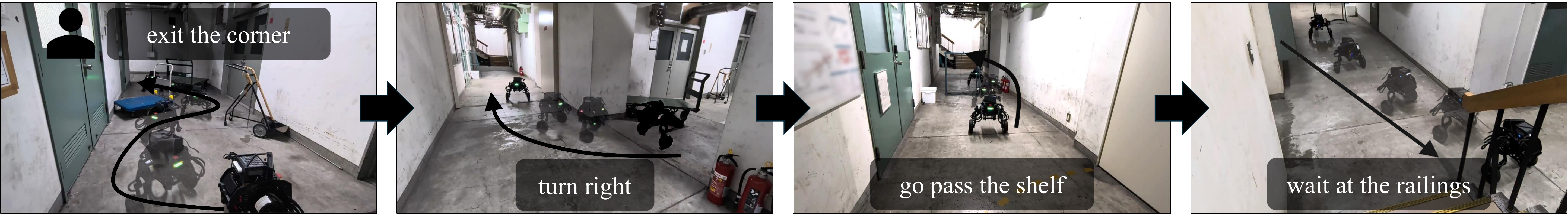}
  \vspace{-1.5em}
  \caption{
    Qualitative snapshots of CoFL-S executing a human-assigned local-instruction sequence in a basement. 
    For each step, we visualize the assigned local instruction.
  }
  \label{fig:realworld_base}
  \vspace{-0.3em}
\end{figure}

\begin{figure}[h]
  \centering
  \includegraphics[width=\linewidth]{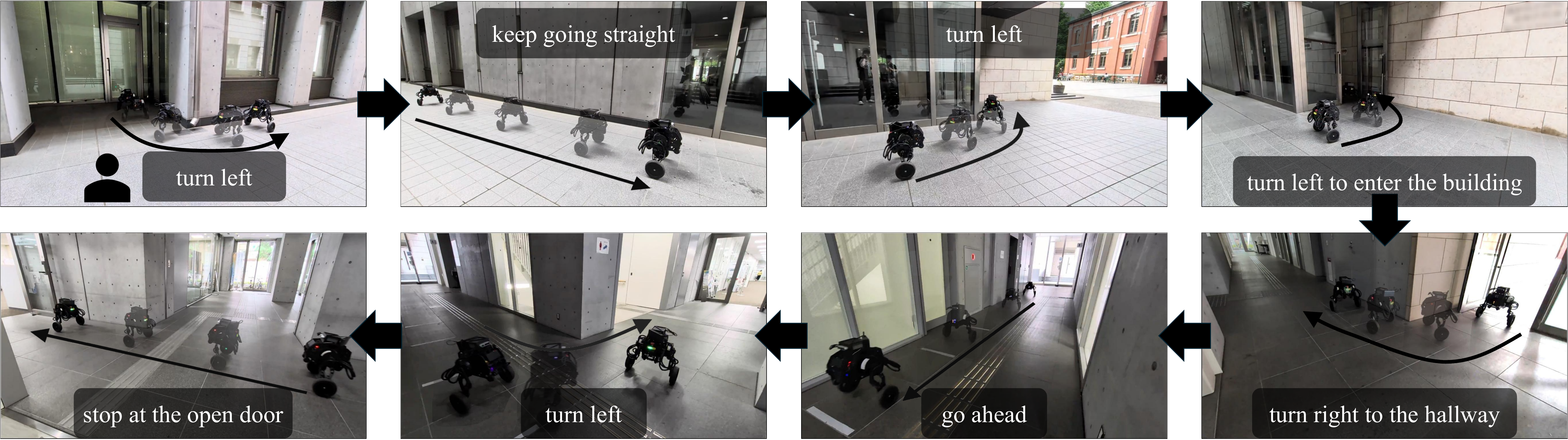}
  \vspace{-1.5em}
  \caption{
    Qualitative snapshots of CoFL-S executing a human-assigned local-instruction sequence in a teaching building. 
    For each step, we visualize the assigned local instruction.
  }
  \label{fig:realworld_tb_1}
  \vspace{-0.3em}
\end{figure}

Additional qualitative snapshots of CoFL-S in real-world experiments are provided in Fig.~\ref{fig:realworld_base} and Fig.~\ref{fig:realworld_tb_1}. 
In these experiments, CoFL-S executes a human-assigned local-instruction sequence, where each local instruction is provided according to the robot's current progress.

\subsubsection{Failure Analysis}
\label{app:failure_analysis}

\begin{wrapfigure}[26]{r}{0.39\linewidth}
  \centering

  \begin{subfigure}{\linewidth}
    \centering
    \includegraphics[width=\linewidth]{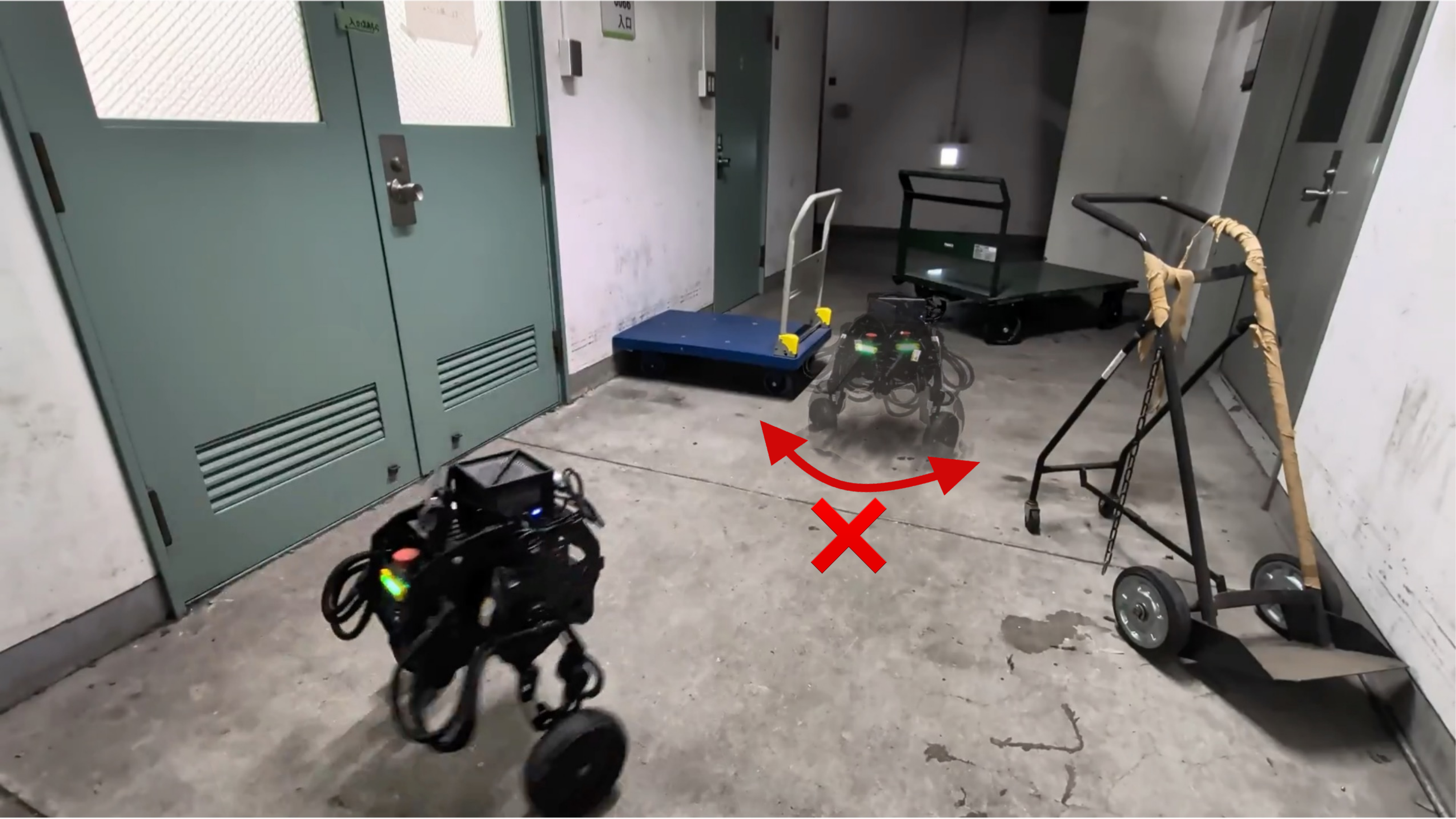}
    \vspace{-1.5em}
    \caption{Typical failure case of Action Token}
    \label{fig:action_token_failure}
  \end{subfigure}
  \par
  \begin{subfigure}{\linewidth}
    \centering
    \includegraphics[width=\linewidth]{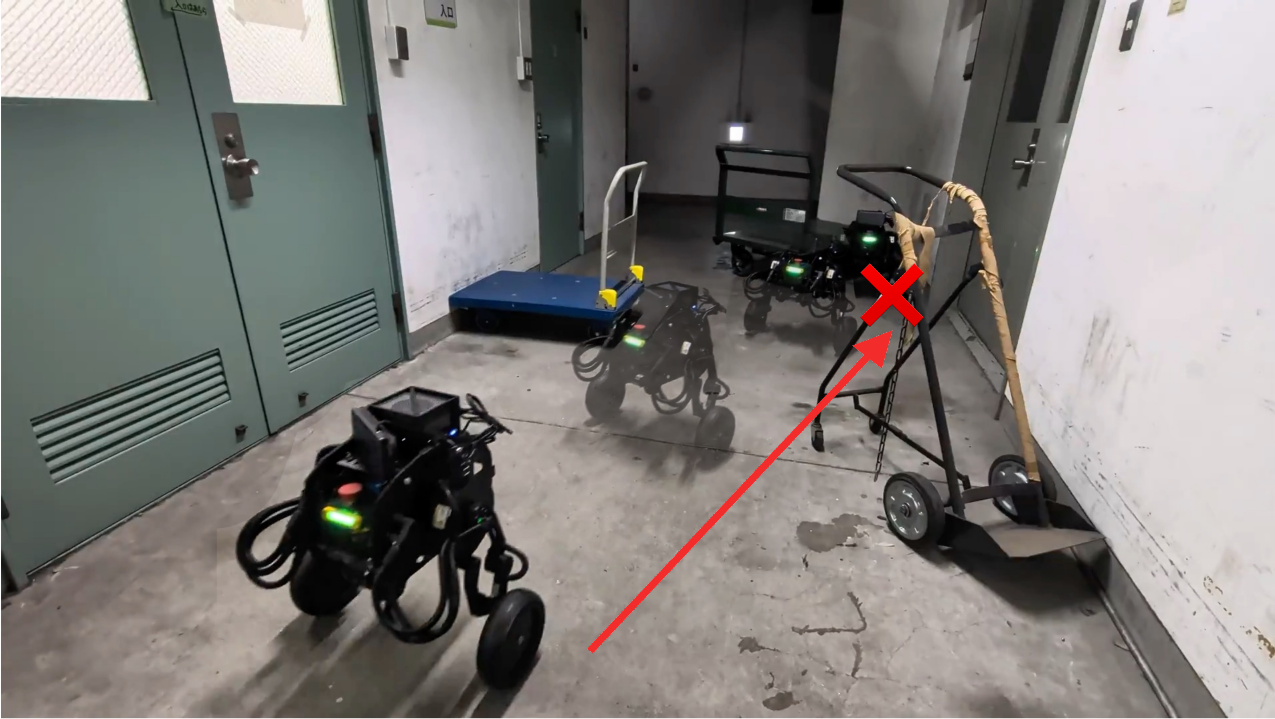}
    \vspace{-1.4em}
    \caption{Typical failure case of Action Chunk}
    \label{fig:action_chunk_failure}
  \end{subfigure}
  \par
  \begin{subfigure}{\linewidth}
    \centering
    \includegraphics[width=\linewidth]{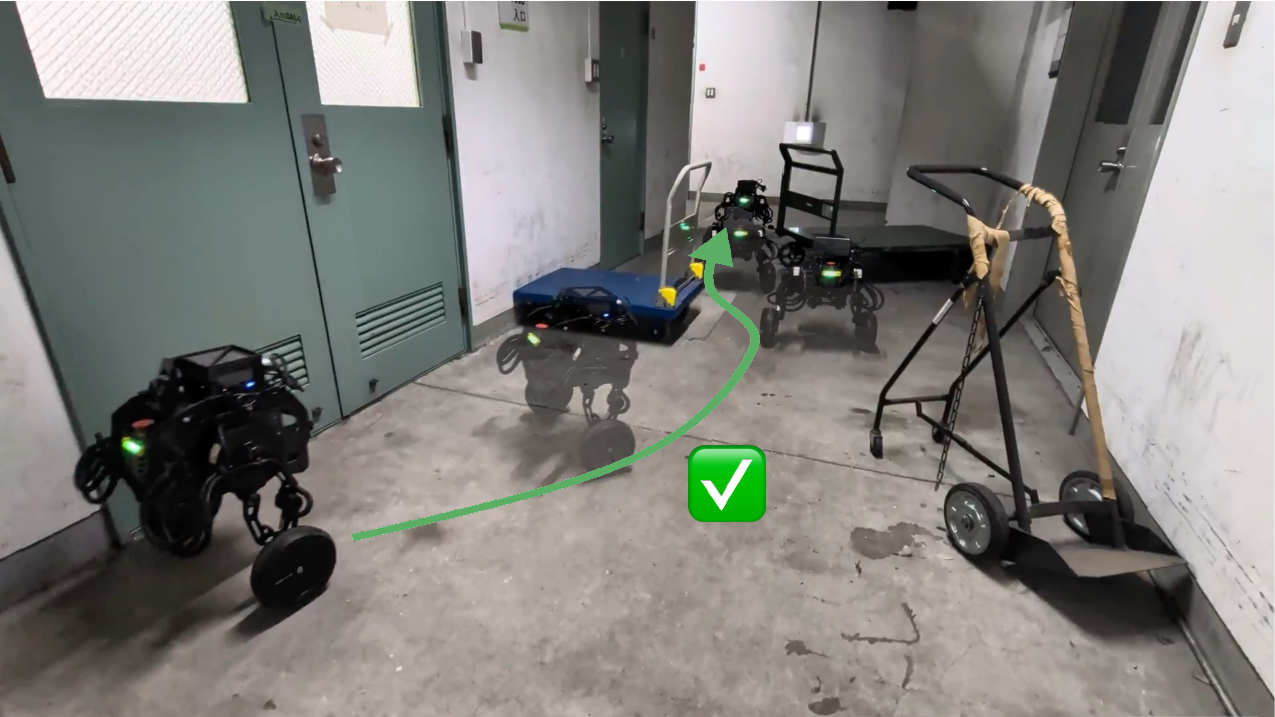}
    \vspace{-1.4em}
    \caption{Successful completion of CoFL-S}
    \label{fig:cofl_success_failure}
  \end{subfigure}

  \vspace{-0.5em}
  \caption{
    Typical real-world outcomes on the same \textit{``exit the corner''} task: Action Token oscillates and gets stuck, Action Chunk drives into a dead end, and CoFL-S exits the corner.
  }
  \label{fig:realworld_failure}
  \vspace{-1.0em}
\end{wrapfigure}

We further analyze representative failure cases observed in the real-world experiments, as shown in Fig.~\ref{fig:realworld_failure}. 

For the \textbf{Action Token} baseline, the most frequent failures occur in narrow passages or near obstacles.
Since the policy predicts only the next discrete motion from the current observation, onboard perception--inference--actuation latency can make the selected action correspond to a stale robot state.
This latency-amplified feedback mismatch leads to repeated left--right corrections rather than stable forward motion, eventually causing the robot to become stuck.

The \textbf{Action Chunk} baseline avoids some of the high-frequency switching behavior by predicting a trajectory. 
However, its predicted trajectory is still anchored to the current observation and does not explicitly encode how the motion should vary across nearby spatial states. 
As a result, it often fails to capture the free space and may generate trajectories that move directly toward a dead end or an obstacle corner. 
This indicates that trajectory-level prediction alone does not necessarily provide sufficient geometric awareness for robust real-world navigation.

In contrast, \textbf{CoFL-S} predicts a spatially queryable sector field and derives motion through field integration.
Because dense sector supervision provides velocity targets at many local spatial queries rather than only along a single trajectory, CoFL-S receives a substantially richer training signal for geometric awareness over the visible workspace. The integrated rollout also provides a short-horizon geometric command, making execution less sensitive to moderate observation--control latency than single-step action prediction.

Nevertheless, all three methods share some failure modes. 
In particular, semantic grounding errors and cases where the target disappears from the ego-centric view are difficult to recover from, even for CoFL-S. 
This suggests that the sector-field interface improves local geometric control, but does not by itself solve long-horizon memory, target re-identification, or semantic recovery.

\end{document}